%% file: DiffusionEncoder.tex
\documentclass{article}

\input{math_commands.tex}

\PassOptionsToPackage{numbers, compress}{natbib}

\usepackage[main, preprint]{neurips_2026}

\usepackage[utf8]{inputenc} 
\usepackage[T1]{fontenc}    
\usepackage[backref=page]{hyperref} 
\usepackage{url}            
\usepackage{booktabs}       
\usepackage{amsfonts}       
\usepackage{nicefrac}       
\usepackage{microtype}      
\usepackage{algorithm}
\usepackage{algorithmic}

\usepackage[svgnames,dvipsnames,x11names,table]{xcolor}
\usepackage{graphicx}
\usepackage{subcaption}
\usepackage{cleveref}
\usepackage{tikz}
\usepackage{pgfplots}
\usetikzlibrary{decorations.markings}
\usetikzlibrary{shapes.misc}
\usetikzlibrary{calc}
\usetikzlibrary{arrows}

\hypersetup{
    colorlinks=true,
    linkcolor={red!50!black},
    citecolor={blue!50!black},
    urlcolor={blue!80!black}
}

\crefname{table}{Table}{Tables}
\crefname{equation}{Eq.}{Eqs.}
\crefname{appendix}{App.}{Apps.}
\crefname{section}{Sec.}{Secs.}
\crefname{figure}{Fig.}{Figs.}
\crefname{algorithm}{Algorithm}{Algorithms}
\crefname{ALC@unique}{step}{steps}

\def \d{{\rm d}}
\def \e{{\epsilon}}

\def \hz{{\hat{z}}}

\def \O{{\mathcal O}}
\def \P{{P}}

\def \s{s}
\def \t{t}
\def \u{\mathfrak{u}}
\def \th{{\bm{\theta}}}

\def \tz{{\tilde{\vz}}}
\def \hz{{\hat{\vz}}}

\def \L{{\mathcal L}}
\def \R{{R}}
\def \T{{T}}

\def \hZ{{\hat{\mZ}}}
\def \pZ{\mathfrak{Z}} 
\def \tZ{{\tilde{\mZ}}}

\def \N{{\mathcal N}}

\def \d{{\textrm{d}}}
\def \e{{\bm e}} 

\def \s{{s}}

\def \t{{t}}
\def \ta{\tilde{\va}}

\def \u{{\mathfrak{u}}}
\def \th{{\bm{\theta}}}
\def \ph{{\bm{\phi}}}
\def \ps{{\bm{\psi}}}
\def \g{{\sigma}}

\def \B{{\mB}} 
\def \KL{{D_\textrm{KL}}}
\def \N{{\mathcal{N}}}

\def \P{{p}}
\def \T{{T}}

\def \DimX{{D_{\mX}}}

\def \DimZ{{D_{\mZ}}}
\def \DimKey{{D_{\rm k}}}
\def \DimVal{{D_{\rm v}}}
\def \Attn{{\textrm{Attention}}}
\def \feat{{\textrm{feat}}}

\def \finP {{\P_{\d}}}

\def \deq {{\, := \,}}
\def \rdeq {{\,=:\,}}
\def \ss {\scriptstyle}

\def \pd{{\partial}}

\newcommand{\AP}[1]{{}}
\newcommand{\SL}[1]{{}}

\title{The Diffusion Encoder}

\author{
  Akhil Premkumar \\
  Department of Physics \\
  University of California San Diego \\
  La Jolla, CA 92093 \\
  \texttt{akhilprem@ucsd.edu} \\
  \And
  Sarah Lucioni\\
  Independent Researcher \\
  \texttt{sklucioni@gmail.com} \\
}

\begin{document}

\maketitle

\begin{abstract}
We construct a new kind of encoder, leveraging the expressive power of diffusion models. In a traditional variational autoencoder, the encoder and decoder jointly negotiate a latent representation of the input. This is made possible by the reparameterization trick, which simplifies training at the cost of restricting the encoder to a simple family of distributions. Replacing this encoder with a diffusion model requires rethinking how the decoder pressure can be transmitted back to the encoder, given that they tend to update their internal estimates of the latent in opposing directions. We solve this problem with an alternating training scheme, inspired by the expectation-maximization algorithm. Our method enables more reliable synchronization between encoder and decoder, while preserving the simple and efficient training objective of standard diffusion models.
\end{abstract}

\section{Introduction}
\label{sec:Introduction}

Can a diffusion model be used as an encoder? After all, diffusion models are highly expressive probabilistic models, which makes them a natural candidate for parameterizing complex distributions over latent variables. The challenge is that these models are typically applied to cases where the target distribution is fixed, whereas a latent distribution must necessarily evolve during training as the encoder and decoder negotiate an optimal representation of the input signal.

In variational autoencoders (VAEs) \cite{Kingma13}, the encoder is a Gaussian distribution with a diagonal covariance. Although this choice limits representational flexibility, it enables joint training of the encoder and decoder through the reparameterization trick. That is, it is straightforward to express the latents in terms of the encoder parameters, which allows the latter to be directly coupled to the decoder pressure. The same is not true for a diffusion model, since the latents are generated through many more iterations over the underlying score network, which makes their functional dependence on the network parameters more complicated \cite{Celik25}. Hence, we need to resort to a training algorithm that alternates between encoder and decoder updates, similar to techniques commonly used in maximum entropy reinforcement learning \cite{Haarnoja18,Li26}.

Sacrificing the convenience of joint training may still be worthwhile if it yields a more flexible latent representation. The limited expressiveness of a rigid encoder family can hinder the autoencoder's effectiveness in the following ways: (1) the optimal encoder may not lie within the chosen family, (2) an overly smooth encoder may map semantically distinct inputs to overlapping neighborhoods in latent space, and (3) an inexpressive encoder can throttle the reconstruction fidelity of the decoder \cite{Cremer18}. These effects are discussed in greater detail in \cref{sec:IBandVAEs}.

Part of the expressive power of a diffusion model comes from its iterative generative process, which allows the model to gradually build up complex internal correlations in the generated sample (see \cref{sec:StochasticEncoder}). Normalizing flows \cite{Rezende15} and neural ODEs \cite{Chen18} also share this feature, but diffusion models benefit from a considerably simpler and more robust training objective (see \cref{sec:TheDiffusionEncoder}). Flow matching models \cite{Lipman23} also inherit this property. However, diffusion models remain differentiated by their firm grounding in information theory, specifically, their ability to efficiently estimate quantities like mutual information and neural entropy \cite{Franzese24,Kong23,Kong24,Premkumar25a}, which are otherwise difficult to compute in high-dimensions. This is particularly relevant to autoencoders, which are, at their core, realizations of an information bottleneck \cite{Tishby2000,Shamir2010, Alemi18}.

In this paper, we develop a framework to leverage a diffusion model as an encoder. The key innovation is a training scheme that updates the encoder and decoder in alternation, while keeping them synchronized with one another (see \cref{sec:EquilibratingToPosterior}). This procedure is directly inspired by the expectation-maximization algorithm \cite{Dempster77}, where the E-step updates a diffusion model with samples from the posterior, and the M-step updates the decoder. The posterior samples are generated from a short Langevin chain that explores the decoder landscape, while an affine drift term keeps the samples regularized. Lifting the regularization pressure from the encoder side subdues the tendency for the encoder and decoder to fall out of alignment with one another during their independent updates. The diffusion encoder takes on a more passive role in this arrangement, which means it can be trained with the standard, well-established denoising objective.
\begin{figure}
    \centering
    \includegraphics[width=\linewidth]{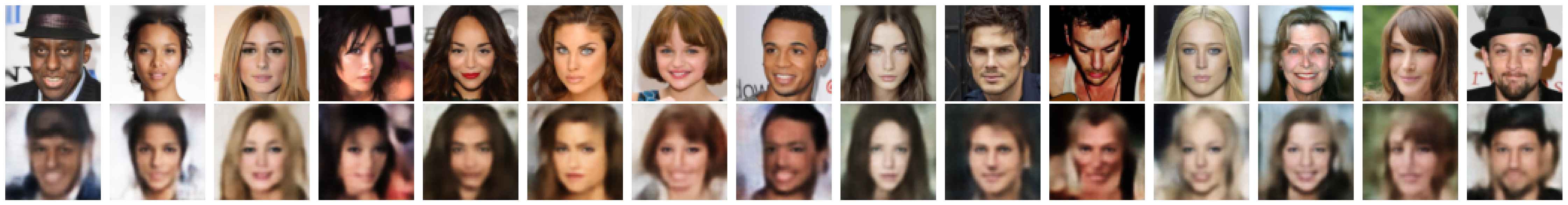}
    \caption{\label{fig:CelebA-recons-small} Input/reconstructed images (top/bottom) from a diffusion encoder + convolutional decoder.}
\end{figure}

Finally, it is important to be clear about the scope of this work. Our goal is not to surpass the state-of-the-art in encoder design; over a decade of refinement has made the Gaussian encoder a highly competitive baseline. Rather, we seek to motivate a new application of diffusion models that draws on their representational capacity and training simplicity \cite{Ho20,Dhariwal2021,Song21}. Therefore, the experiments in \cref{sec:Experiments} serve to demonstrate proof of concept---showing that the proposed setup trains stably and produces coherent reconstructions---rather than optimizing reconstruction quality at scale. We will be transparent about the limitations of our framework as they arise, and suggest potential avenues for improvement throughout. We hope these ideas spur further research on the diffusion encoder.



\section{Information Bottlenecks and Variational Autoencoders}
\label{sec:IBandVAEs}

We begin with an information-theoretic view of unsupervised representation learning, starting from the information bottleneck principle \cite{Tishby2000,Shamir2010}. Consider two random variables $\mX$ and $\mY$ that are correlated with one another (see \cref{sec:Notation} for notation). We want to predict $\mY$ given some value of $\mX$. This can be done by extracting from $\mX$ only the information that is relevant to $\mY$. Such information can be encoded in a third random variable $\mZ$, called the \textit{latent}, which compresses from $\mX$ the details pertinent to $\mY$. The optimal assignment from $\mX$ to $\mZ$ is obtained by finding the encoder $q(\vz|\vx)$ that solves (with $\gamma > 0$)
\begin{equation}
    \min_{q(\vz|\vx)} I(\mX; \mZ) - \gamma^{-1} I(\mZ; \mY) . \label{eq:IB}
\end{equation}
The mutual information (MI) between two random variables quantifies how much we learn about the value of one of them, given a measurement of the other (see App.\ A of \cite{Premkumar26a} for a review). Minimizing $I(\mX; \mZ)$ while maximizing $I(\mZ; \mY)$ squeezes into $\mZ$ only the $\mY$-specific content of $\mX$.

Implementing \cref{eq:IB} in practice requires estimating the MI terms, which is challenging when the random variables are high-dimensional. One workaround is to choose a simple form of the encoder,
\begin{equation}
    q_\ph(\vz | \vx) = \N(\vz; \mu_\ph(\vx), \textrm{diag}(\sigma_\ph(\vx)^2)) , \label{eq:GaussianEncoder}
\end{equation}
and the objective
\begin{equation}
    \min_{\ph, \ps} \E_{\mX, \mY} \E_{q_\ph(\vz | \vx)}[-\log \P_\ps(\vy | \vz)] + \gamma \E_\mX \KL(q_\ph(\vz | \vx) \,\|\, p(\vz))
    \rdeq \min_{\ph, \ps} \E_{\mX, \mY} \L_{\rm VI} . \label{eq:VAE}
\end{equation}
The first term is the \textit{reconstruction} term, which also trains the decoder $\P_\ps(\vx | \vz)$. It follows from
\begin{equation}
    \min_{\ph, \ps} - I(\mZ; \mY)
        = - S(\mY) + \min_{\ph, \ps} S(\mY|\mZ)
        \equiv \min_{\ph, \ps} \E_{\mY, \mZ}[- \log \P_\ps(\vy | \vz) ] ,
        \rdeq \min_{\ph, \ps} \L_{\rm rec}
        \label{eq:ReconstructionTerm}
\end{equation}
since the entropy of $\mY$ is fixed by the dataset, and is independent of the parameters $(\ph, \ps)$.
If we choose $\mY = \mX$, we recognize \cref{eq:VAE} as the negative of the evidence lower bound (ELBO) of a variational \textit{auto}encoder \cite{Kingma13} (see \cref{sec:VAE}). The second term in \cref{eq:VAE}, called the \textit{regularization} term, incentivizes the encoder to be as uninformative as possible, by pushing $q_\ph(\vz | \vx)$ to be as close to the prior $p(\vz) \equiv \N(0,I)$. It is straightforward to show that 
\begin{equation}
    \L_{\rm reg} \deq
    \E_\mX \KL(q_\ph(\vz | \vx) \,\|\, p(\vz))
        = I(\mZ; \mX) + \KL (q_\ph(\vz) \| p(\vz)) .
    \label{eq:RegularizationDecomposition}
\end{equation}
Intuitively, the aggregate $q_\ph(\vz) \deq \E_\mX[q_\ph(\vz | \vx)]$ is the distribution of all available `latent codes,' and $q_\ph(\vz|\vx)$ is concentrated on the subset that encodes a given $\vx$. Minimizing $\L_{\rm reg}$ is equivalent to (1) $\min I(\mZ; \mX)$, which forces $\mZ$ to retain as little information about $\mX$ as possible, and (2) $\min \KL (q_\ph(\vz) \| p(\vz))$, which makes $q_\ph(\vz)$ as close as possible to $p(\vz)$, constraining `total channel capacity.' Up to this additional KL term, \cref{eq:VAE} realizes the bottleneck from \cref{eq:IB}.

The regularization term, $\L_{\rm reg}$, forces $q_\ph(\vz|\vx) \to p(\vz)$ for every $\vx$. In the limit, the encoder ignores the input $\vx$ entirely, leading to posterior collapse \cite{Lucas19,He19}. This can happen, for instance, when the decoder is very powerful; if the decoder can model $\mX$ without any help from $\mZ$, there is little incentive for the encoder to produce informative latents, and the regularization pressure wins.

In a conventional VAE, the KL between $q_\ph(\vz | \vx)$ and $p(\vz)$ is easy to compute. The form of $q_\ph(\vz | \vx)$ in \cref{eq:GaussianEncoder} produces a simple expression for this term,
\begin{equation}
    \KL (q_\ph(\vz | \vx) \,\|\, p(\vz)) =
        \frac{1}{2} \sum_{j=1}^{\DimZ} ((\mu_\ph^j)^2 + (\sigma_\ph^j)^2 - \log (\sigma_\ph^j)^2 - 1) ,
    \label{eq:RegularizationTermVAE}
\end{equation}
where $\mu_\ph^j$ and $\sigma_\ph^j$ are the $j^{\rm th}$ components of $\mu_\ph$ and $\sigma_\ph$. During training, $\vz$'s are generated from the data using $\vz = \mu_\ph(\vx) + \sigma_\ph(\vx) \odot \bm{\epsilon}$, where $\bm{\epsilon} \sim \N(0, I)$. This is called the reparameterization trick, and it allows the gradients to flow through the sampling step, since each $\vz$ is a function of $\ph$. Despite these advantages, the Gaussian encoder from \cref{eq:GaussianEncoder} has some major drawbacks. First, it is not, in general, the optimum of \cref{eq:VAE}. Varying $\L_{\rm VI}$ w.r.t.\ $q_\ph(\vz | \vx)$, with $\mY = \mX$, we find
\begin{equation}
    0 = \frac{\delta \L_{\rm VI}}{\delta q_\ph(\vz | \vx)}
    \implies
    q^\star_\ph(\vz | \vx) \propto p(\vz) e^{\frac{1}{\gamma} \log p_\ps(\vx | \vz)} .
    \label{eq:OptimalEncoderVAE}
\end{equation}
So the optimal encoder is generically non-Gaussian. For $\gamma=1$, we recover the standard result that $q^\star_\ph(\vz | \vx) = p_\ps(\vz | \vx)$, the true posterior. An ideal family of variational distributions $q_\ph(\vz | \vx)$ should be flexible enough to contain \cref{eq:OptimalEncoderVAE} as one solution. Second, the encoder parameters $\ph$ are shared across all inputs, so the encoder cannot optimize independently for each data point $\vx$. Therefore, even if there exists some $\hat{q}_\ph(\vz|\vx)$ within the chosen encoder family that is as close as possible to the ideal in \cref{eq:OptimalEncoderVAE}, the learned encoder $q_\ph(\vz | \vx)$ may be different from $\hat{q}_\ph(\vz|\vx)$ for each individual $\vx$. The total inference gap can therefore be decomposed into an \textit{approximation gap}, and the \textit{amortization gap}, the latter quantifying the inability of the encoder to attain per-instance optimum due to the shared parameter mapping $\vx \to \ph$.
It is shown in \cite{Cremer18,Kim18} that the amortization gap is often the dominant component.
Therefore, the encoder architecture must also support a mapping from $\vx \to q_\ph(\vz|\vx)$ that is expressive enough to specialize to each input, while generalizing reliably to unseen data.

A third, subtler limitation arises from the coupling between the encoder and decoder during training. Since the optimal encoder in \cref{eq:OptimalEncoderVAE} depends on $p_\ps(\vx | \vz)$, the decoder can adjust itself to make $q^\star_\ph(\vz | \vx)$ easier to represent within the chosen encoder family. Thus, the decoder effectively absorbs the encoder's limitations rather than driving the encoder toward a better posterior approximation---co-adapting to a limited encoder ultimately stifles the decoder \cite{Cremer18}.

\begin{figure}
    \centering
    \includegraphics[width=\linewidth]{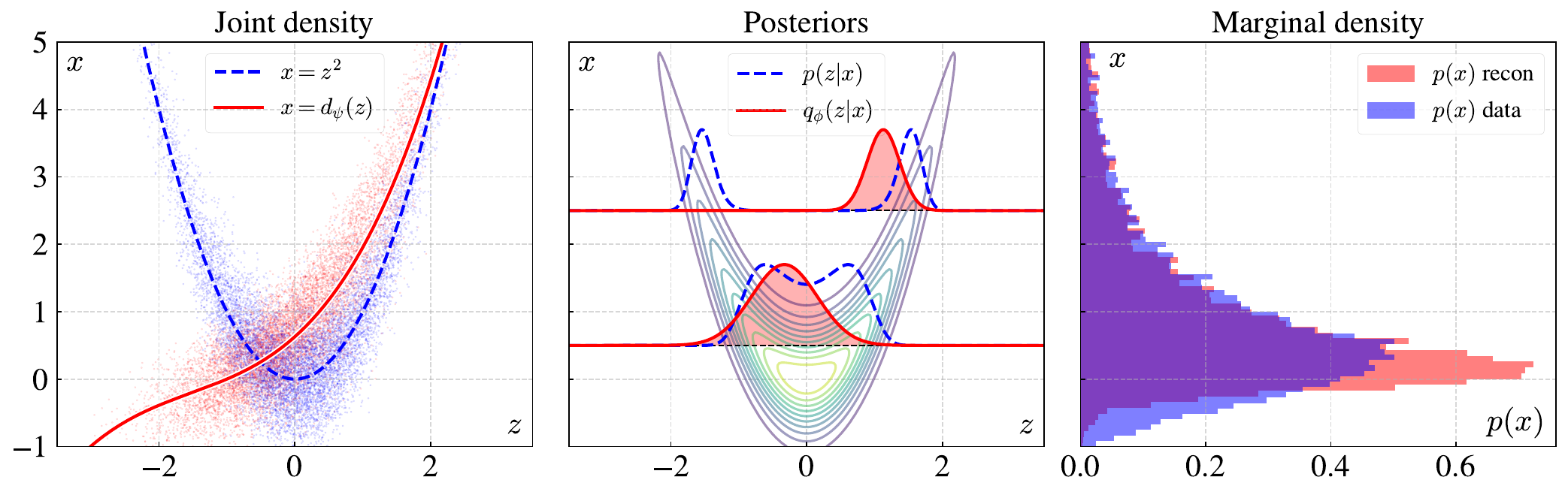}
    \caption{\label{fig:ToyModelVAE} A toy model illustrating the limitations of a Gaussian encoder (cf.\ \cref{eq:ToyPosterior}). \textbf{Left:} Samples from the true joint density $p(x,z)$ (blue) and the learned one (red). \textbf{Middle:} The true posterior at two values of $x$, and the Gaussian encoder that best fits it. \textbf{Right:} The true marginal density and the one reconstructed by the VAE. The Gaussian encoder is unable to capture the true posterior at any $x$, which restricts the decoder to be a mostly linear function of $z$. See also \cref{fig:ToyModeldAE}.}
\end{figure}

\paragraph{A toy example} It is instructive to consider a simple example that illustrates the above concerns. We consider sample generated by a process $z \sim \N(0,1), x|z \sim \N(z^2, \sigma^2)$, where $Z$ and $X$ are one-dimensional random variables. The posterior is
\begin{equation}
    p(z|x) \propto \exp \left( -\frac{(z^2-x)^2}{2 \sigma^2} - \frac{z^2}{2} \right) ,
    \label{eq:ToyPosterior}
\end{equation}
which is \textit{not} a Gaussian in $z$. For large $x$, the posterior is bimodal, with symmetric peaks at $z = \pm \sqrt{x}$. For small $x$, the posterior is concentrated near zero but is non-Gaussian (see \cref{fig:ToyModelVAE}). A Gaussian encoder cannot capture the true posterior at any $x$, which creates an approximation gap.

Next, we consider a simple polynomial decoder $d_\ps(z) = \sum_{k=0}^{3} \psi_{k} z^k$, with learnable coefficients $\ps \equiv \{ \psi_k \}$, and likelihood $p_\ps(x|z) = \N(x; d_\ps(z), \sigma^2)$ (cf.\ \cref{eq:GaussianLikelihood}). The true posterior corresponds to $\psi_2 = 1$ and $\psi_{k}=0$ for all $\ k \neq 2$. However, if we try to approximate the posterior with a Gaussian family, the decoder coefficients $\psi_0$ and $\psi_1$ will be strongest since a linear decoder, $d_\ps(z) = \psi_0 + \psi_1 z$, gives a Gaussian posterior. This is an example of the co-adaptation problem: even though the decoder can express the true solution, the limitations of the encoder prevent it from doing so.


\section{Towards a Stochastic Encoder}
\label{sec:StochasticEncoder}

A simple way of creating a more expressive encoder is to construct a chain of Gaussian encoders, each with its own mean and variance. That is, consider a chain of latent variables $\{ \vz_0, \vz_1, \dots, \vz_n \}$ and the product $q(\vz_n | \vz_{n-1}, \vx) \cdots q(\vz_1 | \vz_0, \vx) q(\vz_0)$ where
\begin{equation}
    q(\vz_{i+1} | \vz_{i}, \vx) = \N(\vz_{i+1}; \mu_\ph(\vz_{i}, \vx), \g_i^2) ,
\end{equation}
and $q(\vz_0) = \N(0,I)$. We have kept the variance $\g_i^2$ of each kernel independent of $\vz_i$ or $\vx$ for simplicity. To see how this setup creates correlations within $\vz_n$, assume for a moment that $\mu_\ph(\vz_i, \vx)^{a} \propto \vz^{a}_{i} + \beta \vz^{b}_{i}$ where $a \neq b$. That is, in the course of producing a sample $\vz_{i+1}$, the kernel mixes into the $a^{\rm th}$ component of $\vz_i$ a term proportional to its $b^{\rm th}$ component. This establishes a correlation between different components, which can grow richer as we travel deeper along the chain. The resulting $\vz_n$ has richer statistics than those afforded by a single Gaussian encoder.

We can stack together a large number of such kernels, provided that the transformation induced by each kernel is scaled appropriately,
\begin{equation}
    q(\vz_{i+1} | \vz_{i}, \vx) = \N(\vz_{i+1}; \vz_{i} + f_\th(\vz_{i}, \vx, \t_i) \Delta \t, \g(t_i)^2 \Delta \t) .
\end{equation}
Here $\t_i \deq i \Delta \t$, and $n \Delta \t \rdeq \T$ is a finite number. The noise variance scales as $\Delta t$, so that the accumulated standard deviation of $\vz_n$ remains $\O(\sqrt{T})$ and stays finite for large $n$. Thus $f_\th$ and $\g$ need not vanish in the continuum limit $n \to \infty$.
The samples from the above kernel are
\begin{equation}
    \vz_{i+1} = \vz_{i} + f_\th(\vz_{i}, \vx, \t_i) \Delta \t + \g(t_i) \sqrt{\Delta t} \bm{\epsilon}_i ,
\end{equation}
where $\bm{\epsilon}_i \sim \N(0, I)$. In the continuum limit, this becomes the stochastic differential equation (SDE)
\begin{equation}
    \d \mZ_\t = f_\th(\mZ_\t, \vx, \t) \d \t + \g(\t) \d \mB_\t . \label{eq:EncoderSDE}
\end{equation}
where $\mB_\t$ is the Wiener process and $\mZ_\t$ is a random variable, a sample of which is $\vz_\t$. Starting from a trivially distributed $\mZ_0$, we can transform to a highly-structured distribution $q_\th(\vz_\T | \vx)$ by applying \cref{eq:EncoderSDE} over time $\T$. The requirement that $q_\th(\vz_\T | \vx)$ remains close to the standard normal $p(\vz)$ is enforced by penalizing deviations of \cref{eq:EncoderSDE} from the SDE
\begin{equation}
    \d \mZ_\t = f_{\rm eq}(\mZ_\t, \t) \d \t + \g(\t) \d \mB_\t . \label{eq:EqSDE}
\end{equation}
where $f_{\rm eq}(\mZ_\t, \t) = - \frac{\g(\t)^2}{2} \mZ_\t$, so $p(\vz)$ is the equilibrium distribution of \cref{eq:EqSDE}. The KL between the path measures $\mathbb{Q}$ and $\mathbb{P}$ generated by \cref{eq:EqSDE,eq:EncoderSDE} upper bounds the KL between the terminal distributions at $\T$, which follows from the data processing inequality (cf.\ Theorem 1 in \cite{Durkan21}):
\begin{equation}
    \KL(\mathbb{Q} \| \mathbb{P}) \geq \KL(q_\th(\vz_\T | \vx) \| p(\vz_\T)) . \label{eq:StochasticRegularization}
\end{equation}
The KL on the left is
\begin{equation}
    \KL(\mathbb{Q} \| \mathbb{P})
        = \E_{\mathbb{Q}} \left[ \int_{0}^{\T} \d \t
                \frac{1}{2 \g(t)^2} \| f_\th(\vz_\t, \vx, \t) - f_{\rm eq}(\vz_\t, \t) \|^2
            \right]
        \label{eq:PathKL}
\end{equation}
This KL would replace the regularization term in the autoencoder objective in \cref{eq:VAE}. During training the encoder generates a sample by, say, an Euler-Maruyama integration of \cref{eq:EncoderSDE}:
\begin{equation}
    \vz_{n} = \vz_{0} + \Delta \t \sum_{i=0}^{n-1} f_\th(\vz_i, \vx, t_i) + \sqrt{\Delta t} \sum_{i=0}^{n-1} \g(t_i) \bm{\epsilon}_i . \label{eq:SolutionEM}
\end{equation}
In principle, we can feed this into the decoder, which then applies pressure on $f_\th$ to shape $\vz_n$ to be maximally informative about $\vx$, while \cref{eq:PathKL} tries to do the opposite. This is the SDE version of the reparameterization trick, and it is used in \cite{Celik25}. But this approach has a few drawbacks: (1) $f_\th$ determines $\vz_\t$ via \cref{eq:SolutionEM}, and we evaluate the $f_\th$ in the loss \cref{eq:PathKL} at those $\vz_\t$, so changing $f_\th$ changes where we evaluate $f_\th$, making the loss landscape non-stationary, and (2) we must store $n$ intermediate states $\vz_1, \dots, \vz_n$ and the activations in $f_\th$ at each step, as well as the noise samples $\bm{\epsilon}_i$, so the memory scales linearly in $n$.

The adjoint method \cite{Chen18,Li20} is a principled solution to these issues. Instead of backpropagating through \cref{eq:SolutionEM}, they solve an adjoint equation backwards in time, which can be used to directly compute the loss gradient w.r.t.\ $\th$. This approach is not without its own difficulties, however, as gradient-based optimization on the resulting non-convex problem has been reported to be empirically unstable \cite{Domingo24}. Adjoint \textit{matching} \cite{Domingo25} resolves this issue by recasting the training objective into a convex form, while also simplifying the backward evolution of the adjoint state. This method is also particularly well-suited to modeling tilted distributions like the one in \cref{eq:OptimalEncoderVAE}.

\paragraph{Fitting a moving target} A fundamental feature of the autoencoder setup is that $q_\th(\vz_\T \| \vx)$ evolves during training. In a classic VAE, the reparameterization trick allows the latent to be expressed as a function of the encoder parameters, so the encoder and decoder can be trained jointly---the decoder influence backpropagates to the encoder parameters via
\begin{equation}
    \nabla_\ph \L_{\rm rec} = \nabla_\vz \L_{\rm rec} \frac{\pd \vz}{\pd \ph} ,
    \qquad
    \vz = \mu_\ph(\vx) + \sigma_\ph(\vx) \odot \bm{\epsilon} .
    \label{eq:DecoderInfluence}
\end{equation}
The analogue for a diffusion encoder would be to let the decoder influence the encoder parameters through \cref{eq:SolutionEM}, but this is computationally expensive and potentially unstable.

On the other hand, adjoint methods are built on the assumption that the target distribution is fixed. During training, samples $\vz_\T$ are generated with \cref{eq:SolutionEM}, which are then used to compute the initial values of the adjoint states at $\t=\T$.
Notably, only the value of $\vz_\T$ is used in this approach, and $f_\th$ is never differentiated through the generative process itself. The parameters $\th$ receive gradient information exclusively via the adjoint equation, which propagates corrections backward in time independent of how $\vz_\T$ was generated.

In the autoencoder context, the adjoint method can be adapted to train the stochastic encoder on a transient target $q_\th(\vz_\T | \vx)$, provided the distribution evolves slowly over the course of training. First, the encoder generates latents $\vz_\T$ with the current $\th$. These $\vz_\T$ are then used to update the encoder and decoder \textit{separately}. This creates a synchronization problem: the encoder and decoder have opposing preferences for what the latent should be, as manifest from the competing terms in \cref{eq:IB}, and updating them in isolation allows this disagreement to compound. A more detailed account of this issue is given in \cref{sec:RegularizationOnEncoderSide}. The setup can still be made to work, albeit with delicate tuning of the training parameters, but our goal is a more robust approach that does not depend on such sensitivity.

\section{Equilibrating to the Posterior}
\label{sec:EquilibratingToPosterior}

Alternating optimization schemes have a storied history in machine learning, with the expectation-maximization (EM) algorithm being one of the oldest and most well-established. We draw inspiration from this method to develop a stable training routine with the stochastic encoder from \cref{sec:StochasticEncoder}.
Briefly, the EM algorithm \cite{Dempster77} is a general method for maximum likelihood estimation in latent variable models. The goal is to maximize the marginal log likelihood $p_\ps(\vx)$, which is exactly
\begin{equation}
    \log p_\ps(\vx) =
        \underbrace{
            \E_{q(\vz | \vx)}[\log \P_\ps(\vx | \vz)] - \KL(q(\vz | \vx) \,\|\, p(\vz))
            }_{\L(q; \ps)}
        + \KL(q(\vz|\vx) \| p_\ps(\vz|\vx)) .
\end{equation}
%
Since the KL term is non-negative, $\L(q; \ps)$ is a lower bound on $\log p_\ps(\vx)$---it is the ELBO (cf.\ \cref{eq:ELBO}). EM finds a solution to the maximum likelihood problem by alternating between two updates, called the E-step and M-step. The expectation step, or E-step, sets $q(\vz | \vx) = p_\ps(\vz | \vx)$ at the \textit{current} values of the parameter $\ps$. Then, the M-steps maximizes $\L(q; \ps)$ over $\ps$ with $q$ fixed. Since the KL term in $\L(q; \ps)$ is independent of $\ps$, this reduces to
\begin{equation}
    \max_\ps \L(q; \ps) = \max \E_{q(\vz | \vx)}[\log \P_\ps(\vx | \vz)] .
\end{equation}
The bound is tight at the start of each M-step, since setting $q$ to the true posterior makes $\log \P_\ps(\vx) = \L(q; \ps)$. Therefore, any increase in $\L(q; \ps)$ guarantees a corresponding increase in $\log p_\ps(\vx)$, ensuring monotone convergence. We can adopt the same strategy for the stochastic encoder.

\begin{figure}
    \centering
    \includegraphics[width=\linewidth]{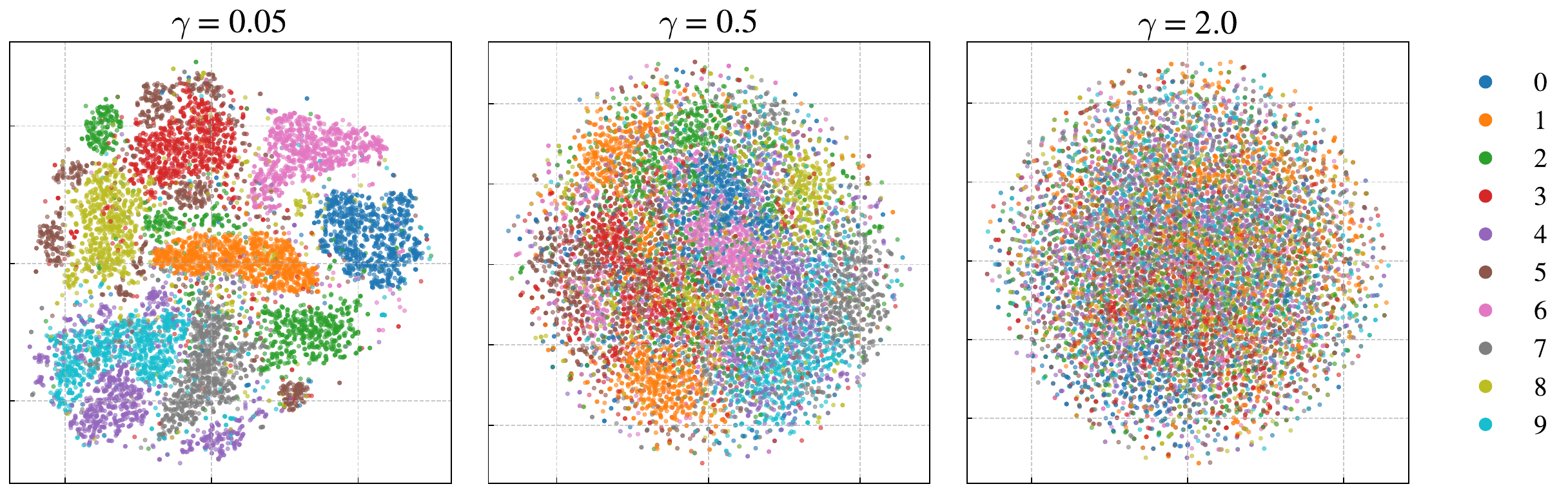}
    \caption{\label{fig:MNIST_tSNE_ldim20} t-SNE diagram of latents (MNIST, $\DimZ=20$) learned by the diffusion encoder. Each figure shows the equilibrium distribution $q^\star(\vz|\vx)$ of \cref{eq:ELBO,eq:PosteriorSDE}, at a given temperature $\gamma$.  As $\gamma$ increases, stochasticity dominates in \cref{eq:PosteriorSDE}, scrambling the latent clusters into the prior $p(\vz)$.}
\end{figure}

We propose a novel scheme, where the encoder directly models the optimal solution from \cref{eq:OptimalEncoderVAE},
\begin{equation}
    q^\star(\vz | \vx)
        \propto p(\vz) e^{\frac{1}{\gamma} \log p_\ps(\vx | \vz)}
        \propto \exp \left( -\frac{\vz^2}{2} + \frac{1}{\gamma} \log p_\ps(\vx|\vz) \right) .
    \label{eq:OptimalEncoder}
\end{equation}
Notice that this is the equilibrium distribution of an SDE (with $\ps$ fixed),
%
\begin{equation}
    \d \mZ_\tau = \left[ -\mZ_\tau + \frac{1}{\gamma} \nabla \log p_\ps(\mX|\mZ_\tau) \right] \d \tau + \sqrt{2} \, \d \mB_\tau .
    \label{eq:PosteriorSDE}
\end{equation}
%
%
With sufficient iteration, \cref{eq:PosteriorSDE} converges to samples $\vz_\star \sim q^\star(\vz | \vx)$, which can then be used to train a diffusion model that will serve as the stochastic encoder (see \cref{fig:MNIST_tSNE_ldim20,sec:TheDiffusionEncoder}). This is the E-step. The Langevin evolution makes the latents generated from \cref{eq:PosteriorSDE} consistent with the current decoder, even if they do not start out that way. More importantly, the encoder is no longer responsible for regularization---this is handled by the affine term in the drift---relegating it to a more passive role in which any tendency for misalignment is subdued. This addresses the synchronization problem. Next, the reconstruction term can be optimized on the same $\vz^\star$, which is the M-step.  Iterating these two steps yields \cref{alg:TrainingDiffEncAE}.
\begin{algorithm}
    \caption{Training the autoencoder}
    \label{alg:TrainingDiffEncAE}
    \begin{algorithmic}[1]
    \REQUIRE Dataset $\gX$, number of Langevin steps $n_{\rm step}$, step size $\Delta \tau$
    \REPEAT
        \STATE Sample minibatch $\{\vx\} \sim \gX$
        \vspace{0.5em}
        \STATE \textit{// E-step: update posterior with equilibrium samples}
        \STATE Generate $\{\vz\} \sim q_\th(\vz_\T | \vx)$ \label{stp:Encode}
        \FOR{$k = 1, \ldots, n_{\rm step}$} \label{stp:BeginLangevin}
            \STATE $\vz \leftarrow \vz + \left[ -\vz + \frac{1}{\gamma} \nabla_\vz \log p_\ps(\vx | \vz) \right] \Delta \tau + \sqrt{2 \Delta \tau}\, \bm{\epsilon}, \quad \bm{\epsilon} \sim \mathcal{N}(0, I)$
            \label{stp:Equilibrate}
        \ENDFOR \label{stp:FinishLangevin}
        \STATE Set $\vz^\star \leftarrow \vz$ \hfill \textit{// posterior samples} \label{stp:CopyResult}
        \vspace{0.5em}
        \STATE Update $\th$ by minimizing the denoising loss with targets $\{\vz^\star\}$ \label{stp:EncoderUpdate}
        \STATE \textit{// M-step: update decoder}
        \STATE Update $\ps$ by minimizing $\E_{\mX} \E_{\mZ^\star|\mX} \left[-\log p_\ps(\vx | \vz^\star)\right]$ \label{stp:Mstep}
    \UNTIL{converged}
    \end{algorithmic}
\end{algorithm}

At first glance, the Langevin iteration appears expensive, since an SDE generally requires many steps to equilibrate. However, $n_{\rm step}$ need not be large if the initial latents are in the vicinity of their equilibrium values, which is the case if $p_\ps$ evolves slowly enough across the entire training run. In other words, we should choose $\gamma$, $\Delta \tau$, and the learning rates such that \cref{eq:PosteriorSDE} relaxes to $p_\ps$ within $n_{\rm step}$ iterations, starting from the freshly generated latents in \cref{stp:Encode}. The encoder can afford to lag behind a little, since the Langevin dynamics will pick up the slack as it updates $\vz_\T \to \vz^\star$ in \cref{stp:BeginLangevin,stp:Equilibrate,stp:FinishLangevin,stp:CopyResult}. This gives us more leeway in choosing $\gamma$ and the learning rates, which would otherwise need to be tuned delicately to keep the encoder and decoder synchronized over the entire training run.  Optionally, we can also decrease $n_{\rm steps}$ as training converges, since fewer Langevin steps are needed to equilibrate to a quasi-static decoder landscape (see \cref{sec:Latency}).



On a practical note, we clip $\nabla_\vz \log p_\ps(\vx | \vz)$ to the interval $(-1, 1)^\DimZ$ to ensure that \cref{eq:PosteriorSDE} does not take arbitrarily large steps. Since the clipped gradient preserves the sign of each component, it continues to push $\vz$ towards regions of high posterior probability where clipping is no longer needed. The samples $\vz^\star$ used to train the encoder are drawn from the stationary distribution $q^\star(\vz|\vx)$, which is largely unaffected by the clipping scheme.

\section{The Diffusion Encoder}
\label{sec:TheDiffusionEncoder}

The final component in our autoencoder setup is a flexible probabilistic model capable of capturing $q^\star(\vz|\vx)$ from its samples. We use a diffusion model for the purpose---the titular \textit{diffusion encoder}. Diffusion models convert noise to data by learning to reverse a diffusion process that transforms data to noise. To use such a model in the paradigm described above, the transformation $\mZ_0 \to \mZ_\T|\vx$ under \cref{eq:EncoderSDE} would be replaced by the generative stage of the diffusion model. This is the reversal of a forward process that converts $\mZ_\T|\vx \to \mZ_0$ via
%
\begin{equation}
    \d \tZ_\s = f_{\rm eq}(\tZ_\s, \T-\s) \d \s + \g(\T-\s) \d \hat{\mB}_\s , \label{eq:ForwardSDE}
\end{equation}
where we have introduced the forward time variable $\s \deq \T - \t$, so $\tZ_0 \equiv \mZ_\T$ and $\tZ_\T \equiv \mZ_0$ (see \cref{fig:ReverseDiffusion}). Over a large $\T$, \cref{eq:ForwardSDE} reaches a stationary distribution, which we will also choose as the prior, $q(\tz_\T) \equiv p(\tz_\T)$. The ideal reverse process is
\begin{equation}
    \d \mZ_\t = -\left[ f_{\rm eq}(\mZ_\t, \t)  - \g(\t)^2 \nabla \log q(\mZ_\t, \t | \vx) \right] \d \t + \g(\t) \d \mB_\t . \label{eq:ReverseSDE}
\end{equation}
Here $q(\mZ_\t, \t | \vx) \equiv q(\tZ_\s, \s | \vx)$ is obtained by applying \cref{eq:ForwardSDE} to $\tZ_0|\vx$ for time $\s$. To learn the reverse process we take $f_\th \equiv f_{\rm eq} + \g^2 \e_\th$ to beget the neural-network parameterized SDE 
\begin{equation}
    \d \mZ_\t = \left[ f_{\rm eq}(\mZ_\t, \t) + \g(\t)^2 \e_\th(\mZ_\t, \T-\t; \vx) \right] \d \t + \g(\t) \d \mB_\t .
    \label{eq:EntropyMatchingSDE}
\end{equation}
This is the \textit{entropy-matching} parameterization, introduced in \cite{Premkumar25a}. It makes the calculation of certain information-theoretic quantities more transparent, as we will see in a moment. For now, we note that we can switch to the score-matching parameterization by simply setting $\bm{s}_\th = 2 f_{\rm eq}/\g^2 + \e_\th$. We train \cref{eq:EntropyMatchingSDE} to match \cref{eq:ReverseSDE} by minimizing the KL between their path measures \cite{Durkan21},
\begin{gather}
    \L^{\mZ|\mX}_{\rm EM}
        \deq \E_{\mX} \left[
                \int_{0}^{\T} \d \s \frac{\g^2}{2} \E_{\tZ_\s, \tZ_0|\vx} \left[
                    \big\| \nabla \log p(\tz_\s) - \nabla \log q(\tz_\s, \s | \vx) + \e_\th(\tz_\s, \s; \vx) \big\|^2
                \right]
            \right] 
        \label{eq:LossEM} \\
    \text{where,  }
    \nabla \log p(\tz) = -\tz \overset{(\ref{eq:EqSDE})}{=} \frac{2 f_{\rm eq}}{\g^2} . \label{eq:EqScore}
\end{gather}
%
The expectation is taken over the paths generated by \cref{eq:EqSDE}, which are the same as those produced by \cref{eq:ForwardSDE} as it transforms $\tZ_0|\vx \to \tZ_\T|\vx$. In practice, we only have access to $q(\vz_\T|\vx) \equiv q(\tz_0|\vx)$ through its samples $\{\tz_0\}$. However, given a sufficient number of such samples, we can train a diffusion model without access to the true score $\nabla \log q(\cdot|\vx)$. This is possible because minimizing \cref{eq:LossEM} is equivalent to minimizing the \textit{denoising} objective \cite{Vincent11},
\begin{align}
    &\L^{\mZ|\mX}_{\rm DEM}
        \deq \E_{\mX}
            \left[
                \int_{0}^{\T} \d \s \frac{\g^2}{2} \E_{\tZ_\s, \tZ_0|\vx}
                \big\| \nabla \log p(\tz_\s) - \nabla \log q(\tz_\s, \s | \tz_0) + \e_\th(\tz_\s, \s; \vx) \big\|^2
            \right] \label{eq:LossEMTheoretical} \\
        &\quad \approx
        \T \, \E_{\s \sim \mathcal{U}(\varepsilon, \T]}
        \E_{\begin{subarray}{l}
            \vx \sim \finP(\vx) \\
            \tz_0 \sim q(\tz_0|\vx) \\
            \tz_\s \sim q(\tz_\s, \s | \tz_0)
            \end{subarray}}
        \left[ \lambda(\s) \frac{\g(\s)^2}{2}
                \big\| \nabla \log p(\tz_\s) - \nabla \log q(\tz_\s, \s | \tz_0) + \e_\th(\tz_\s, \s; \vx) \big\|^2
            \right] .
        \label{eq:LossEMDenoising}
\end{align}
In the second step, we have approximated the integral and expectations with a Monte Carlo average. More importantly, the unknown density $q(\tz_\s, \s | \vx)$ has been replaced by the transition kernel, which is the Gaussian \cite{Karras22}
\begin{equation}
    q(\tz_\s, \s | \tz_0) = \N \left( \tz_\s ; \sqrt{\bar{\alpha}(\s)} \tz_0, (1 - \bar{\alpha}(\s)) I \right) ,
    \qquad
    \bar{\alpha}(\s) = \exp \left( - \int_{0}^{s} \d \bar{\s} \, \beta(\bar{\s}) \right) ,
    \label{eq:KernelVP}
\end{equation}
for the VP SDE with $f_{\rm eq}(\tz, \s) = -\frac{\beta(\s)}{2} \tz$ and $\g(\T-\s) = \sqrt{\beta(\s)}$. Thus, the unknown score in \cref{eq:LossEM} is replaced with
\begin{equation}
     \nabla \log q(\tz_\s, \s | \tz_0) = - \frac{\bm{\epsilon}}{\sqrt{1 - \bar{\alpha}(\s)}} ,
     \qquad \bm{\epsilon} \sim \N(0, I) . \label{eq:ScoreKernelVP}
\end{equation}
Notice that the denominator vanishes at $\s=0$, which is why the Monte Carlo estimate starts at $\s = \varepsilon$, a small positive number. We have also introduced a factor $\lambda(\s)$ in going from \cref{eq:LossEMTheoretical} to \cref{eq:LossEMDenoising}, which re-weights the objective in a time-dependent way. The choice $\lambda(\s) = 2(1-\bar{\alpha}(\s))/\beta(\s)$ yields the ``variance dropped'' training objective, which gives better empirical results \cite{Ho20,Durkan21}.

The Gaussian kernel in \cref{eq:KernelVP} also enables \textit{simulation-free} sampling. That is, we can compute averages of the form $\int_\s \E_{\ss \tZ_\s}[\cdots]$ without solving \cref{eq:ForwardSDE} iteratively over each time step. Instead, we can use the kernel to sample any point in a full trajectory in a single jump,
\begin{equation}
    \tz_\s = \sqrt{\bar{\alpha}(\s)} \tz_0 + \sqrt{1 - \bar{\alpha}(\s)} \bm{\epsilon} .
    \label{eq:PropagatedVP}
\end{equation}
This is one of the key features that make diffusion model training so efficient. For comparison, the classical adjoint methods, as well as adjoint matching, lack this property (see \cref{sec:AdjointMatching}). Faster training of the encoder helps offset a portion of the latency due to the Langevin iterations in \cref{alg:TrainingDiffEncAE}.

Finally, setting $f_\th \equiv f_{\rm eq} + \g^2 \e_\th$ in \cref{eq:PathKL}, we see that the KL between the diffusion encoder output and the prior is upper bounded by the conditional \textit{neural entropy} \cite{Premkumar25a,Premkumar26a},
\begin{equation}
    S^{\mZ|\mX}_{\rm NN} \deq
        \E_\mX \left[
            \int_{0}^{\T} \d \s \, \frac{\g^2}{2}
            \E_{\tZ_\s, \tZ_0|\vx} \left[ \left\Vert \e_\th(\tz_\s, \s; \vx) \right\rVert^2 \right]
        \right]
        \geq \KL(q_\th(\vz_\T | \vx) \| p(\vz_\T)) .
    \label{eq:NeuralEntropyXZ}
\end{equation}
Briefly, the neural entropy $S{\ss ^{\mZ|\mX}_{\rm NN}}$ measures the amount of information injected by the diffusion model to drive $q(\vz_0)$ to $q_\th(\vz_\T | \vx)$. It is the information that was captured and stored in the neural network during training. If $S{\ss ^{\mZ|\mX}_{\rm NN}}$ is small the encoder produces a $\mZ_\T$ that is uninformative about $\mX$. This is a satisfying physical picture: $\mZ_\T$ can carry no more information about $\mX$ than the network has stored.

\paragraph{Architecture} Given an input $\vx$, we generate a latent sample from $q_\th(\vz_\T | \vx)$ using the probability-flow ODE, with a Heun solver \cite{Karras22,Maoutsa20}. As the flow evolves $\vz_0 \to \vz_\T$, the network $\e_\th$ extracts information from $\vx$ and injects it into $\vz_\t$ progressively. Therefore, the network architecture must \textit{query} $\vx$ in a $\t$-dependent way. This is a job for attention \cite{Vaswani17}. The details are given in \cref{sec:EncoderArchitecture}.

\section{Experiments}
\label{sec:Experiments}

\begin{figure}
    \centering
    \begin{subfigure}[t]{0.48\linewidth}
        \centering
        \includegraphics[width=\linewidth]{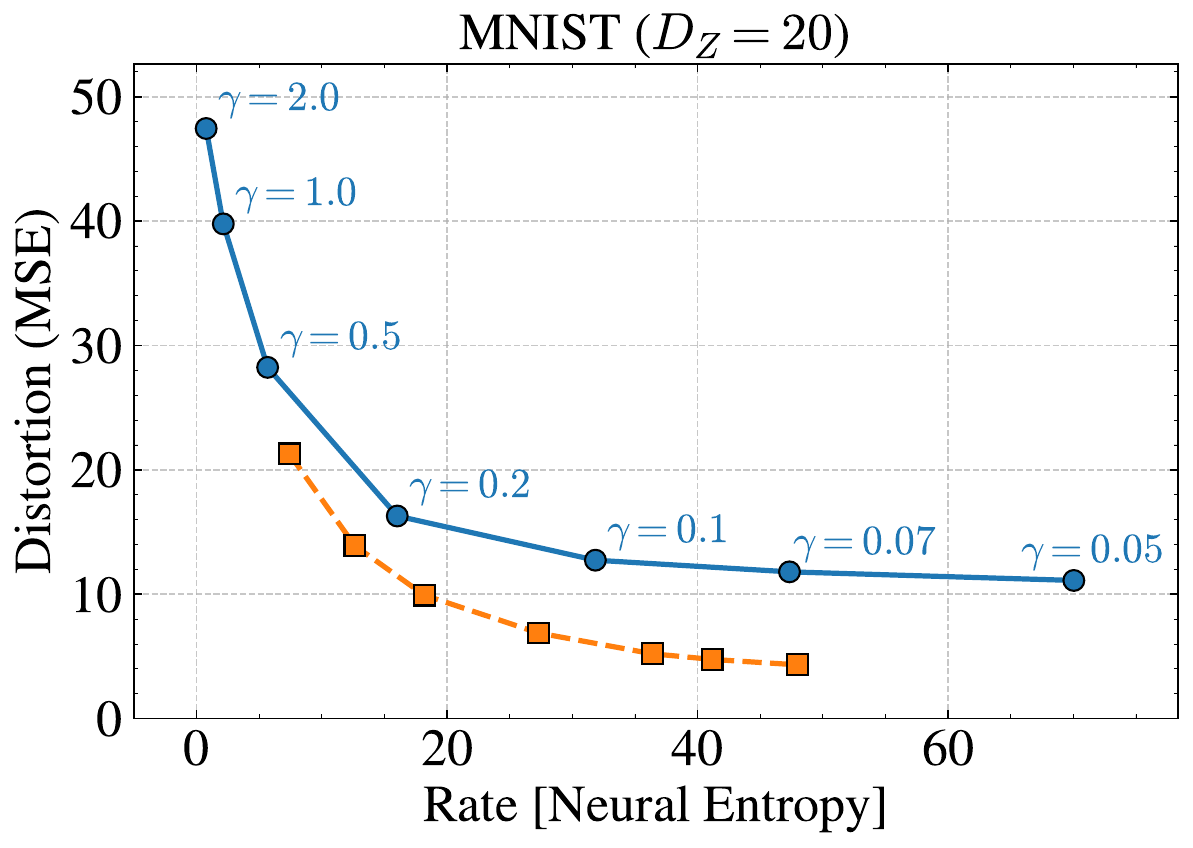}
    \end{subfigure}
    \hfill
    \begin{subfigure}[t]{0.48\linewidth}
        \centering
        \includegraphics[width=\linewidth]{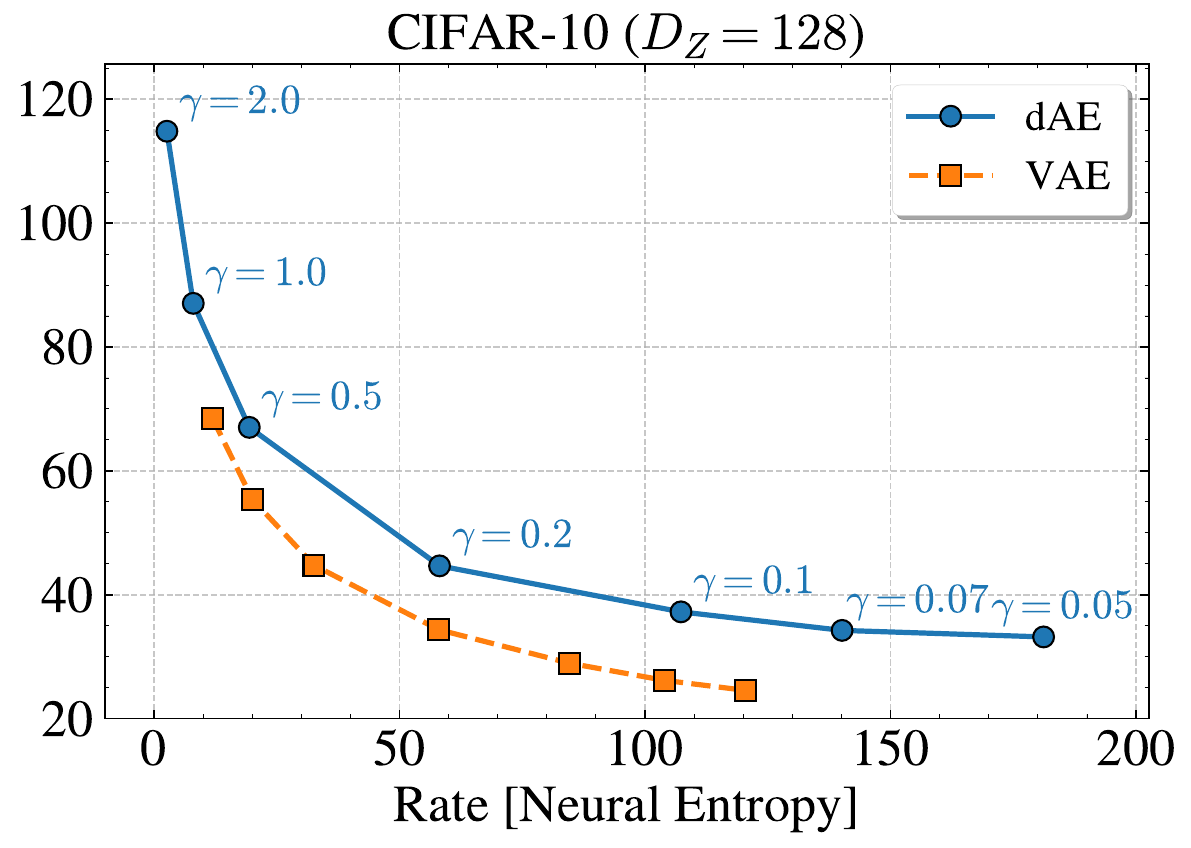}
    \end{subfigure}
    \caption{\label{fig:RateDistVAEvsdAE} Rate-Distortion curves for a VAE and dAE (diffusion encoder + conv.\ decoder).}
\end{figure}

We test our autoencoder on four different image datasets: MNIST, CIFAR-10, TinyImageNet, and CelebA-HQ, the last two scaled to $64 \times 64$ pixels \cite{LeCun98,Krizhevsky09,Deng09,Liu2015,Karras17}. Hyperparameter choices for these experiments are given in \cref{tab:ArchHyperparameters}. Qualitative samples and their reconstructions are provided in \cref{fig:rd_sweep_dAE,fig:rd_sweep_VAE,fig:CelebA-recons,fig:TIN_loss_recons}. We were especially alert to training instabilities, which is precisely what \cref{alg:TrainingDiffEncAE} is designed to counter (see \cref{sec:RegularizationOnEncoderSide}). Empirically, we observed no significant stability issues, even across varying hyperparameter regimes. Moreover, the loss profiles were smooth and nearly monotonic, consistent with an EM-like training scheme (see \cref{fig:TIN_training_loss}). This indicates that the encoder and decoder evolved in lockstep rather than competing with one another. Such synchronization is noteworthy, given that even in a standard VAE, the encoder can lag behind the decoder, leading to posterior collapse \cite{He19}.

Next, we evaluate our autoencoder using rate-distortion (RD) curves, which characterize the fundamental tradeoff between compression and reconstruction accuracy \cite{Alemi18}. For a VAE, rate refers to the regularization term from \cref{eq:RegularizationTermVAE}, and for the diffusion encoder, we use its neural entropy as a close proxy (cf.\ \cref{eq:NeuralEntropyXZ}). Distortion measures the fidelity of the output.  We use a decoder with a standard MSE loss in our experiments, so distortion corresponds to the average squared distance between the input and its reconstruction (cf.\ \cref{eq:DecoderVAE}). In short, RD curves trace the equilibrium between the mutual information terms in the information bottleneck, \cref{eq:IB}, over a range of temperatures $\gamma$.

\cref{fig:RateDistVAEvsdAE} shows the RD curves for a VAE and dAE. The latter refers to our diffusion encoder + convolutional decoder setup. The VAE uses the same decoder, paired with a symmetric convolutional encoder (see \cref{sec:DecoderArchitecture}). RD curves positioned closer to the origin represent superior performance, as they achieve lower reconstruction error at a given rate. Therefore, it is clear from \cref{fig:RateDistVAEvsdAE} that the classic VAE fares better than the dAE. However, it reveals that the dAE operates very close to the VAE's RD frontier, despite having a fundamentally different approach to posterior approximation.

The VAE encoder directly optimizes the quantities being plotted in \cref{fig:RateDistVAEvsdAE}, namely $\L_{\rm rec}$ and $\L_{\rm reg}$ from \cref{eq:ReconstructionTerm,eq:RegularizationTermVAE}. The diffusion encoder learns the optimal solution $q^\star(\vz|\vx)$ from its samples. But the VAE encoder offers better performance, despite the fact that it is a Gaussian (\cref{eq:GaussianEncoder}). To understand this better, we return to the toy model from \cref{sec:IBandVAEs}, and train a diffusion encoder + polynomial decoder on its samples $\{x\}$. We find that the diffusion encoder picks up on the $z \to -z$ symmetry of the posterior, but \textit{still produces a nearly Gaussian $q_\th(z|x)$} even though it can capture the true $p(z|x)$ (see \cref{fig:ToyModeldAE}). We wonder if an analogous statement is true in the general case, and whether that implies some form of \textit{universality} of the Gaussian encoder.

Of course, this is not the only explanation. First, neural entropy only upper-bounds the rate in the diffusion encoder case, so there might be a gap between the quantities we are using on the $x$-axis for the VAE and DAE RD curves. Second, there may be some additional source of error in modeling $q^\star(\vz|\vx)$ from its samples that adds to the distortion. Given the intricacy of the diffusion encoder, it is probable that multiple factors contribute to this discrepancy. We see this as fertile ground for continued research.





\section{Conclusion}
\label{sec:Conclusion}

Can a diffusion model be used as an encoder? Yes, absolutely. Is there more work to be done before it is ready for primetime? Also yes.

The main contribution of our paper is \cref{alg:TrainingDiffEncAE}, which enables (1) synchronized training of a decoupled encoder and decoder pair, and (2) training of the diffusion encoder with the standard denoising objective. A key enabler of our diffusion encoder is the architecture described in \cref{sec:EncoderArchitecture}, which progressively injects information from the input into the latent over the course of the denoising process. We do not claim this to be an optimal design, and expect that more careful architectural choices would yield further improvements.

The primary weakness of \cref{alg:TrainingDiffEncAE} is the sequential overhead introduced by the Langevin chains---patience is the price we pay for stability. Here again, there is room for improvement, by adaptive adjustment of $n_{\rm steps}$ as training settles, and/or the use of critically damped dynamics to reach the stationary distribution more rapidly. We elaborate on these ideas in \cref{sec:Latency}. Fortunately, for lower-dimensional $\mX$ training times remain reasonable; this makes our setup a strong candidate for certain maximum entropy reinforcement learning problems \cite{Haarnoja18,Li26}.

In closing, we note that our training scheme exhibits a certain resilience to posterior collapse, as noted in \cref{sec:Experiments}. This makes sense, since \cref{alg:TrainingDiffEncAE} is explicitly designed to accommodate a certain amount of lag from the encoder, allowing the encoder to catch up with the decoder if it sprints ahead. Whether this structural robustness has broader applications beyond our current setting remains an intriguing open question.

\clearpage
\phantomsection
\addcontentsline{toc}{section}{References}
\bibliographystyle{unsrtnat}
\bibliography{encoder}


\newpage

\addcontentsline{toc}{section}{Contents}
\begingroup
  \tableofcontents
\endgroup
\newpage

\appendix

\section{Outtakes}
\label{sec:Outtakes}

We present some approaches we tried in the course of this work, explaining the rationale behind those ideas, and why they were ultimately scrapped in favor of the method in the main text. This section is intended to save future researchers the effort of rediscovering these detours, and perhaps spark inspiration for more elegant solutions.

\subsection{Regularization on the Encoder side}
\label{sec:RegularizationOnEncoderSide}

A notable feature of \cref{alg:TrainingDiffEncAE} is that regularization happens through Langevin dynamics under \cref{eq:PosteriorSDE}. Equilibrating this SDE is also the highest latency step in the whole process. Another approach would be to use a regularization term analogous to \cref{eq:RegularizationTermVAE}, that throttles the information content of $\mZ$ by directly constraining the encoder's network parameters. For the diffusion encoder, this is accomplished through the conditional neural entropy (cf.\ \cref{eq:NeuralEntropyXZ}),
\begin{equation}
    S^{\mZ|\mX}_{\rm NN} \deq
        \E_\mX \left[
            \int_{0}^{\T} \d \s \, \frac{\g^2}{2}
            \E_{\tZ_\s, \tZ_0|\vx} \left[ \left\Vert \e_\th(\tz_\s, \s; \vx) \right\rVert^2 \right]
        \right]
        \geq \KL(q_\th(\vz_\T | \vx) \| p(\vz_\T)) .
    \label{eq:NeuralEntropyXZAppx}
\end{equation}
Combining $S{\ss ^{\mZ|\mX}_{\rm NN}}$ with the denoising objective (cf.\ \cref{eq:LossEMDenoising}), we obtain
\begin{equation}
    \L_{\rm DEnc} \deq \L^{\mZ|\mX}_{\rm DEM} + \gamma S^{\mZ|\mX}_{\rm NN} .
    \label{eq:DEncRegularizationSingle}
\end{equation}
By adjusting $\gamma$ we can throttle the amount of information stored in the encoder network, which bounds the information compressed into $\mZ_\T$ from $\mX$. Thus, the regularization responsibility shifts back to the encoder. Conveniently, the neural entropy can also be estimated by simulation-free sampling, just like the denoising loss.

The encoder and decoder are still decoupled, so we need to train them with an alternating scheme. This is \cref{alg:TrainingWithNeuralEntropy}. In each cycle, fresh latents are generated by the encoder. We update these to the current decoder parameters, by taking a step (or steps) in the direction of $\nabla_\vz L_{\rm rec}$, where $L_{\rm rec}(\ps, \hz) \deq - \log p_\ps(\vx |\hz)$. Here again, we found that generating $B_\vz$ latent samples per input $\vx$ provided a better training signal to the encoder later, provided we average the updated latents over $B_\vz$ before training the encoder on them (see step 7).
\begin{algorithm}
    \caption{Training the autoencoder with neural entropy}
    \label{alg:TrainingWithNeuralEntropy}
    \begin{algorithmic}[1]
    \REQUIRE Dataset $\gX$, latents per input $B_\vz$, latent step size $\Delta \tau$
    \REPEAT
        \STATE Sample minibatch $\{\vx^{(j)}\}_{j=1}^{B} \sim \gX$
        \vspace{0.5em}
        \FOR{$j = 1, \ldots, B$ and $k = 1, \ldots, B_\vz$}
            \STATE Generate $\hz^{(j,k)} \sim q_\th(\vz_\T | \vx^{(j)})$ by running the PF-ODE with different initial seeds
        \ENDFOR
        \vspace{0.5em}
        \STATE Update latents $\hz^{(j,k)} \leftarrow \hz^{(j,k)} - \Delta \tau \nabla_{\hz} L_{\rm rec} \left( \ps, \hz^{(j,k)} \right)$
        \STATE Compute mean latents $\langle \hz \rangle^{(j)} = \frac{1}{B_\vz} \sum_{k=1}^{B_\vz} \hz^{(j,k)}$
        \STATE Update $\ps$ by minimizing $-\frac{1}{B \cdot B_\vz} \sum_{j,k} \log p_\ps(\vx^{(j)} | \hz^{(j,k)})$
        \vspace{0.5em}
        \STATE \textit{// Encoder update: fit diffusion encoder to mean latents}
        \STATE Update $\th$ by minimizing $\L^{\hZ|\mX}_{\rm DEM} + \gamma S^{\hZ|\mX}_{\rm NN}$ with targets $\{ \langle \hz \rangle^{(j)}, \vx^{(j)} \}_{j=1}^{B}$
    \UNTIL{converged}
    \end{algorithmic}
\end{algorithm}

\paragraph{Synchronization} \cref{alg:TrainingWithNeuralEntropy} works reasonably well on smaller datasets like MNIST and CIFAR-10. However, it suffers from a synchronization problem, like the one described at the end of \cref{sec:StochasticEncoder}. In step 6, the latents are updated based on the \textit{current} state of the decoder. But in step 10, the decoder changes, which means the new latents are no longer `in sync' with the updated decoder parameters. We can make this explicit by adding some time subscripts:
\begin{subequations}
  \begin{gather}
      \hz^{(j)}_{\tau + \Delta \tau} \leftarrow \hz^{(j)}_{\tau} - \Delta \tau \nabla_{\hz} L_{\rm rec} \left( \ps_{\tau}, \hz^{(j)}_{\tau} \right) , \label{eq:UpdateLatentWTime} \\
      \ps_{\tau + \Delta \tau} \leftarrow \ps_{\tau} - \eta_\ps \frac{1}{B} \sum_{k=1}^{B} \nabla_\ps L_{\rm rec} \left( \ps_{\tau}, \hz^{(j)}_{\tau + \Delta \tau} \right) , \label{eq:UpdateDecoderWTime}
  \end{gather}
\end{subequations}
%
%
where we have taken $B_\vz=1$ for simplicity, and $\eta_\ps$ is the decoder learning rate. At the end of one cycle, $\hz{\ss ^{(j)}_{\tau + \Delta \tau}}$ and $\ps{\ss _{\tau + \Delta \tau}}$ are not perfectly consistent with one another. But this is asynchronicity is exacerbated by the fact that the neural entropy term in \cref{eq:DEncRegularizationSingle} dampens the ability of the diffusion encoder to learn $\{ \hz{\ss ^{(j)}_{\tau + \Delta \tau}} \}$ faithfully; its purpose is to \textit{resist} the decoder's preference for what the latents should be. Therefore, the latents generated at the start of the next cycle may be farther away from where the decoder expects them to be. This misalignment can rapidly compound and destabilize training. In practice, we find that \cref{alg:TrainingWithNeuralEntropy} often drives the training loss into a \texttt{nan}.

Despite its shortcomings, \cref{alg:TrainingWithNeuralEntropy} was a crucial stepping stone to \cref{alg:TrainingDiffEncAE}. The crucial realization was that moving the regularization responsibility from the encoder side leads to much better consistency between the latents and decoder during training.

\subsection{Latency}
\label{sec:Latency}

The biggest drawback with \cref{alg:TrainingDiffEncAE} is the latency induced by sequential iteration of the posterior SDE, \cref{eq:PosteriorSDE}, each step of which involves taking a derivative of the decoder network. As mentioned there, $n_{\rm steps}$ does not have to be large for the Langevin chain to equilibrate, since the encoder places the initial latents close to their final destination. We typically set $n_{\rm steps} = 10$ in our experiments (see \cref{tab:ArchHyperparameters}). This parameter is what allows the setup to tolerate some slack between the encoder and decoder, as the latter races ahead in \cref{stp:Mstep}. Even though there is no regularization pressure on the encoder side, the encoder lags because it was last updated with a $\vz_\star$ that may no longer reflect the decoder's preference at the end of each cycle.

The lag is largest early in training, when the decoder loss landscape is changing most rapidly. As the losses stabilize, the encoder catches up and tracks the decoder more closely. It also means we no longer need as many $n_{\rm steps}$ for the latents to settle to $\vz_\star$. In fact, it is possible to adaptively lower $n_{\rm steps}$ as training progresses with little to no effect on synchronization. We did this experiment, but there was a catch: we used JAX as our ML framework \citep{JAX18}, and it requires a separate JIT-compilation of the model for each $n_{\rm steps}$ setting. For smaller datasets, this offsets the latency improvements from adaptive $n_{\rm steps}$, but cleverer ML engineers than us should be able to resolve this.

Another possibility is to introduce a velocity variable into the Langevin dynamics, imbuing $\vz_\tau$ with inertia. This smoothens the evolution of $\vz_\tau$, which (1) regulates it from making large jumps when it encounters regions with large $\nabla_{\vz_\tau} \log p_\ps(\vx|\vz_\tau)$, and (2) leads to faster equilibration if the dynamics is tuned correctly. We did not try this ourselves, but we state the idea here for future development.



\section{More Experiments}
\label{sec:MoreExperiments}

This section collects together a host of experiments that could not be inserted in the main paper due to space constraints. We include more comprehensive RD curves and reconstructions from several experiments.
We trained on H200 GPUs with 141 GB of memory. While most models were trained using four GPUs in parallel, the wall-clock times reported in \cref{tab:ArchHyperparameters} reflect performance on a single GPU to provide a standardized baseline. We used JAX/Flax as our ML framework \cite{JAX18}.
Code will be released upon publication.

\begin{figure}
    \centering
    \begin{subfigure}[t]{0.48\linewidth}
        \centering
        \includegraphics[width=\linewidth]{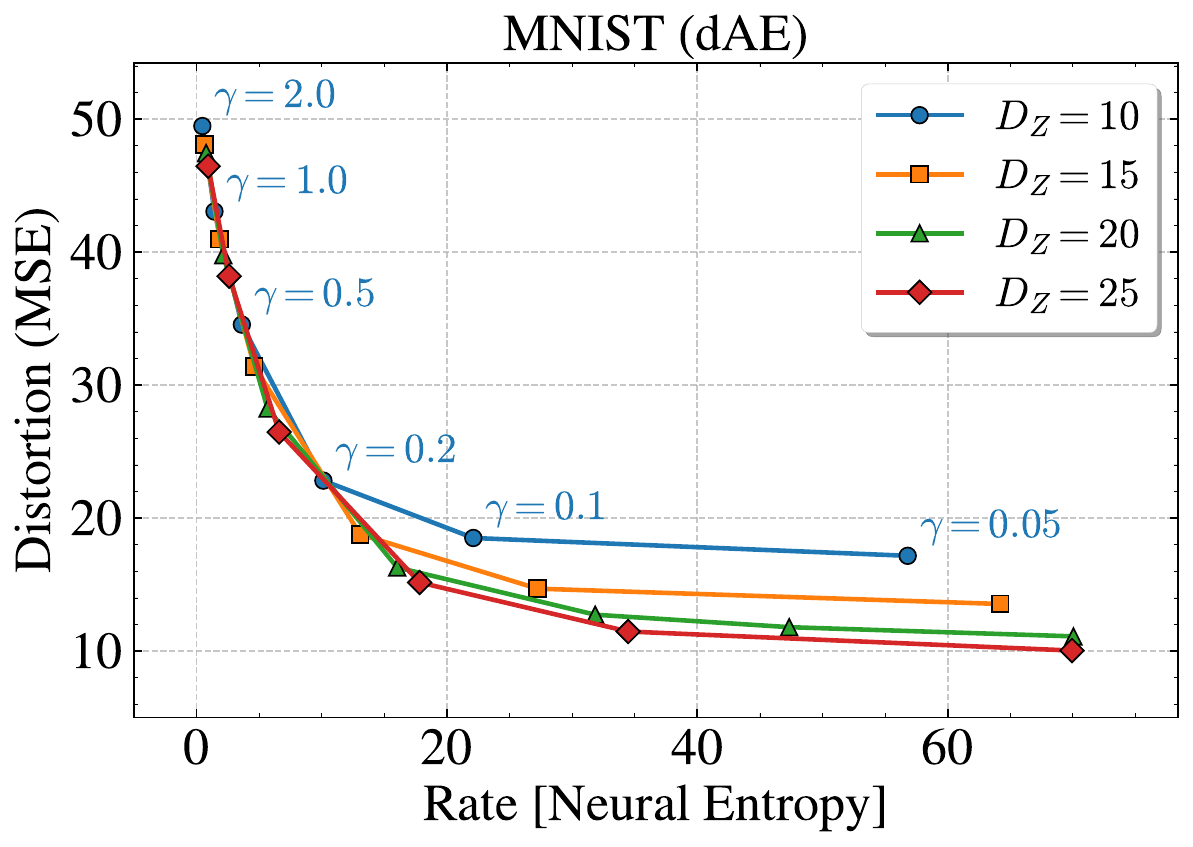}
    \end{subfigure}
    \hfill
    \begin{subfigure}[t]{0.48\linewidth}
        \centering
        \includegraphics[width=\linewidth]{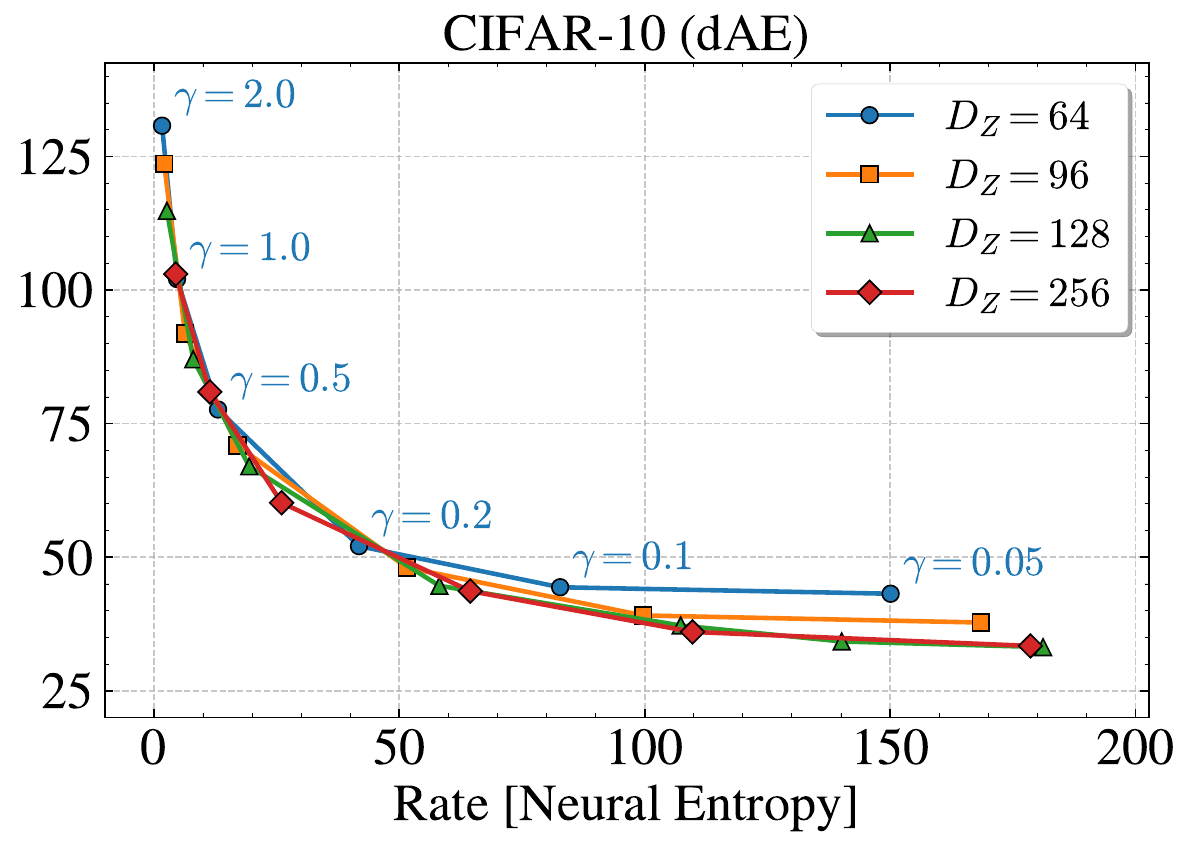}
    \end{subfigure}
    \\[1em]
    \begin{subfigure}[t]{0.83\linewidth}
        \centering
        \includegraphics[width=\linewidth]{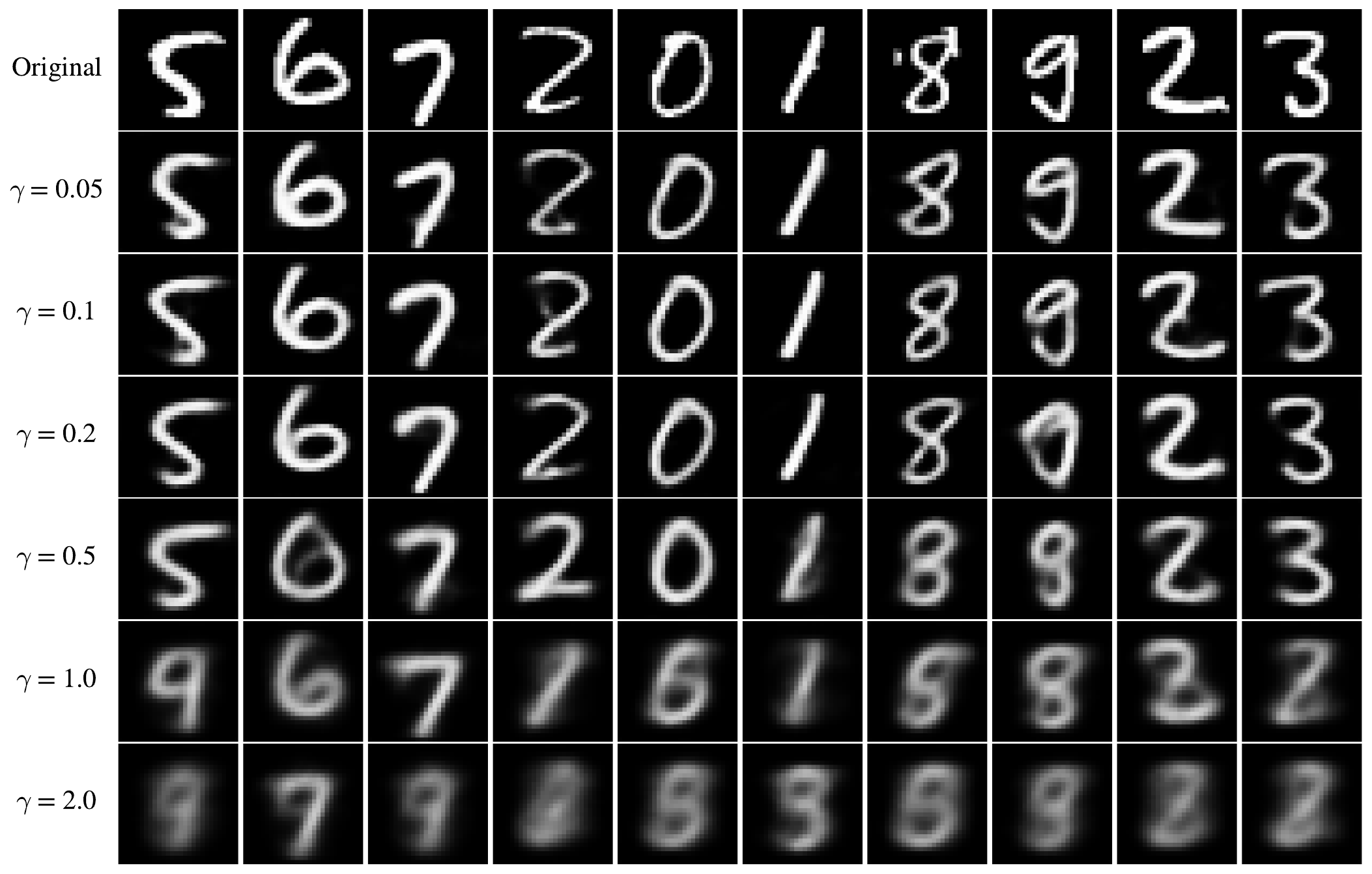}
        \caption{\label{fig:rd_sweep_recons_MNIST} Reconstructions of MNIST samples with a dAE at different temperatures $\gamma$.}
    \end{subfigure}
    \\[1em]
    \begin{subfigure}[t]{0.83\linewidth}
        \centering
        \includegraphics[width=\linewidth]{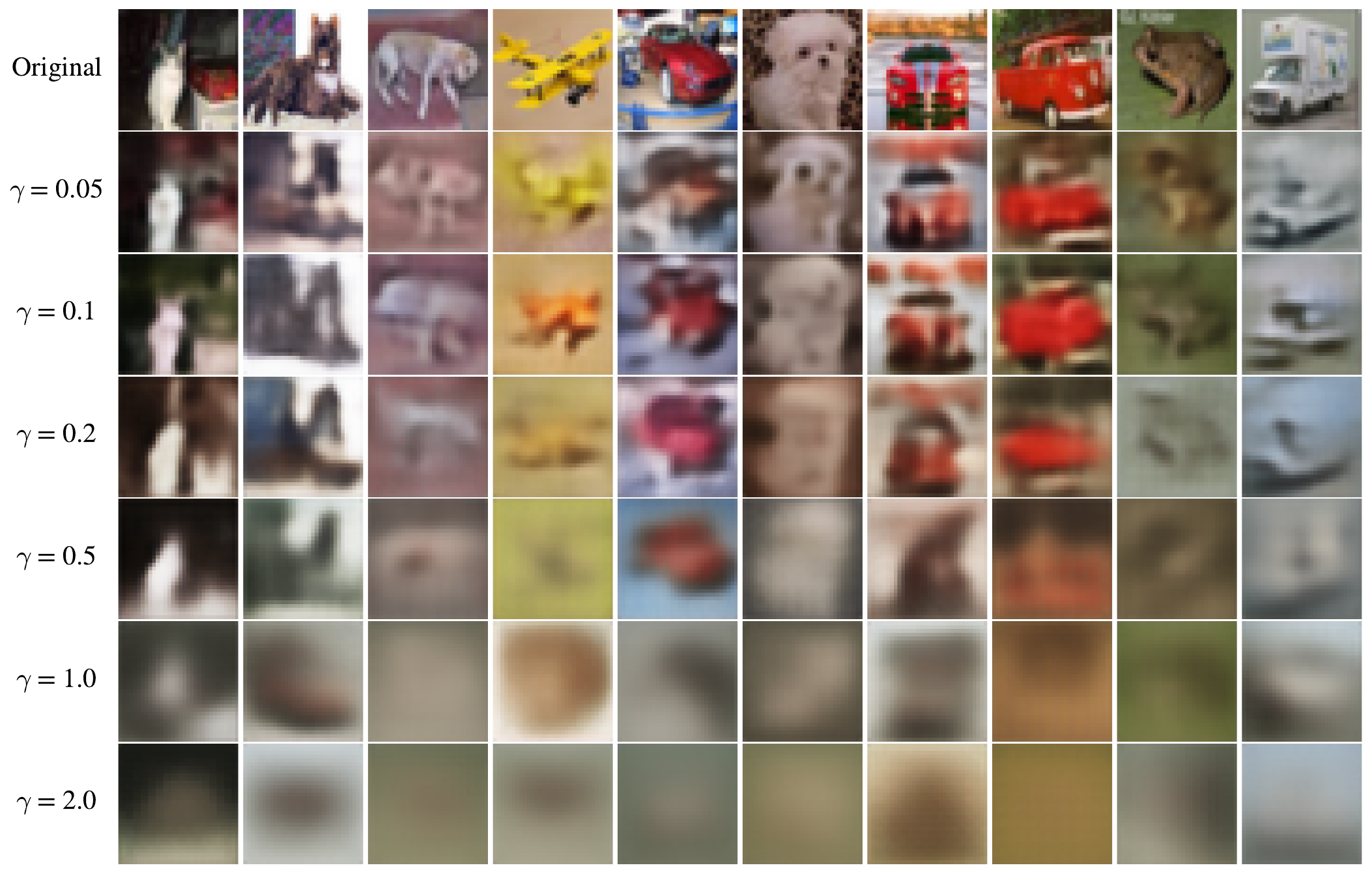}
        \caption{\label{fig:rd_sweep_recons_CIFAR-10} Reconstructions of CIFAR-10 samples with a dAE at different temperatures $\gamma$.}
    \end{subfigure}
    \caption{\label{fig:rd_sweep_dAE} Rate-distortion curves and reconstructions with a diffusion encoder + convolutional decoder (dAE) for MNIST and CIFAR-10. RD curves for different latent dimensions $\DimZ$ are shown. Reconstructions were generated with $\DimZ=20$ for MNIST, and $\DimZ=128$ for CIFAR-10.}
\end{figure}

\begin{figure}
    \centering
    \begin{subfigure}[t]{0.5\linewidth}
        \centering
        \includegraphics[width=\linewidth]{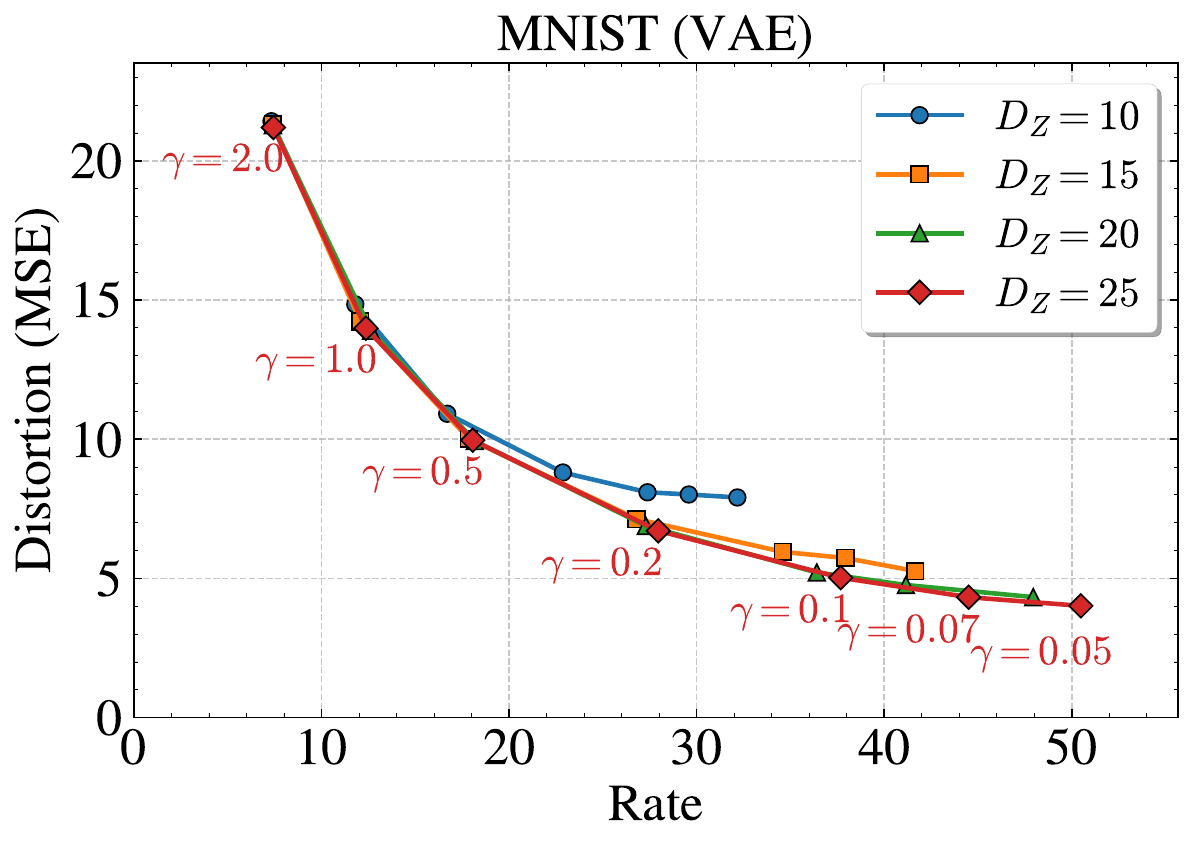}
        \caption{}
        \label{fig:rd_sweep_MNIST_VAE}
    \end{subfigure}
    \hfill
    \begin{subfigure}[t]{0.48\linewidth}
        \centering
        \includegraphics[width=\linewidth]{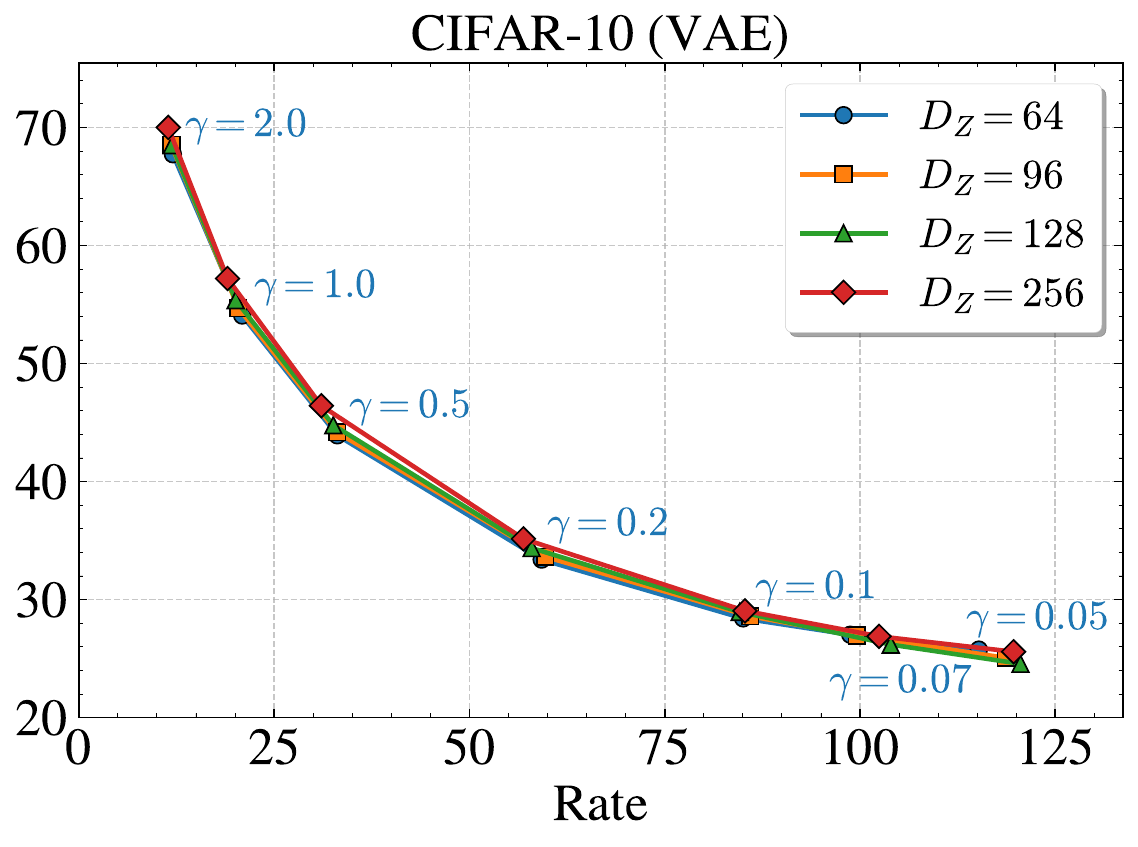}
        \caption{}
        \label{fig:rd_sweep_CIFAR10_VAE}
    \end{subfigure}
    \vspace{1.0em}
    \begin{subfigure}[t]{0.7\linewidth}
        \centering
        \includegraphics[width=\linewidth]{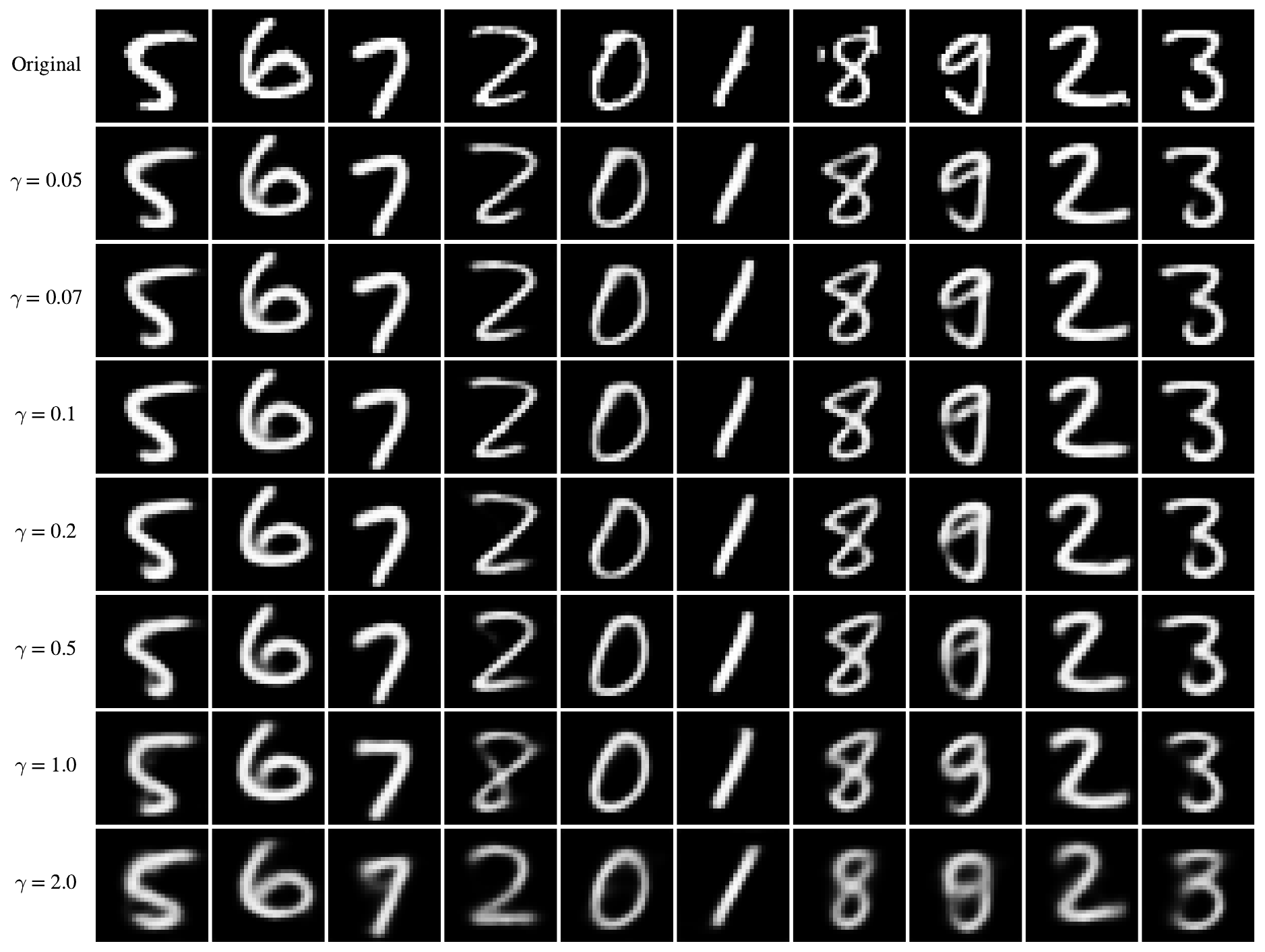}
        \caption{\label{fig:rd_sweep_recons_MNIST_VAE} Reconstructions of MNIST samples with a VAE at different $\gamma$.}
    \end{subfigure}
    \begin{subfigure}[t]{0.7\linewidth}
        \centering
        \includegraphics[width=\linewidth]{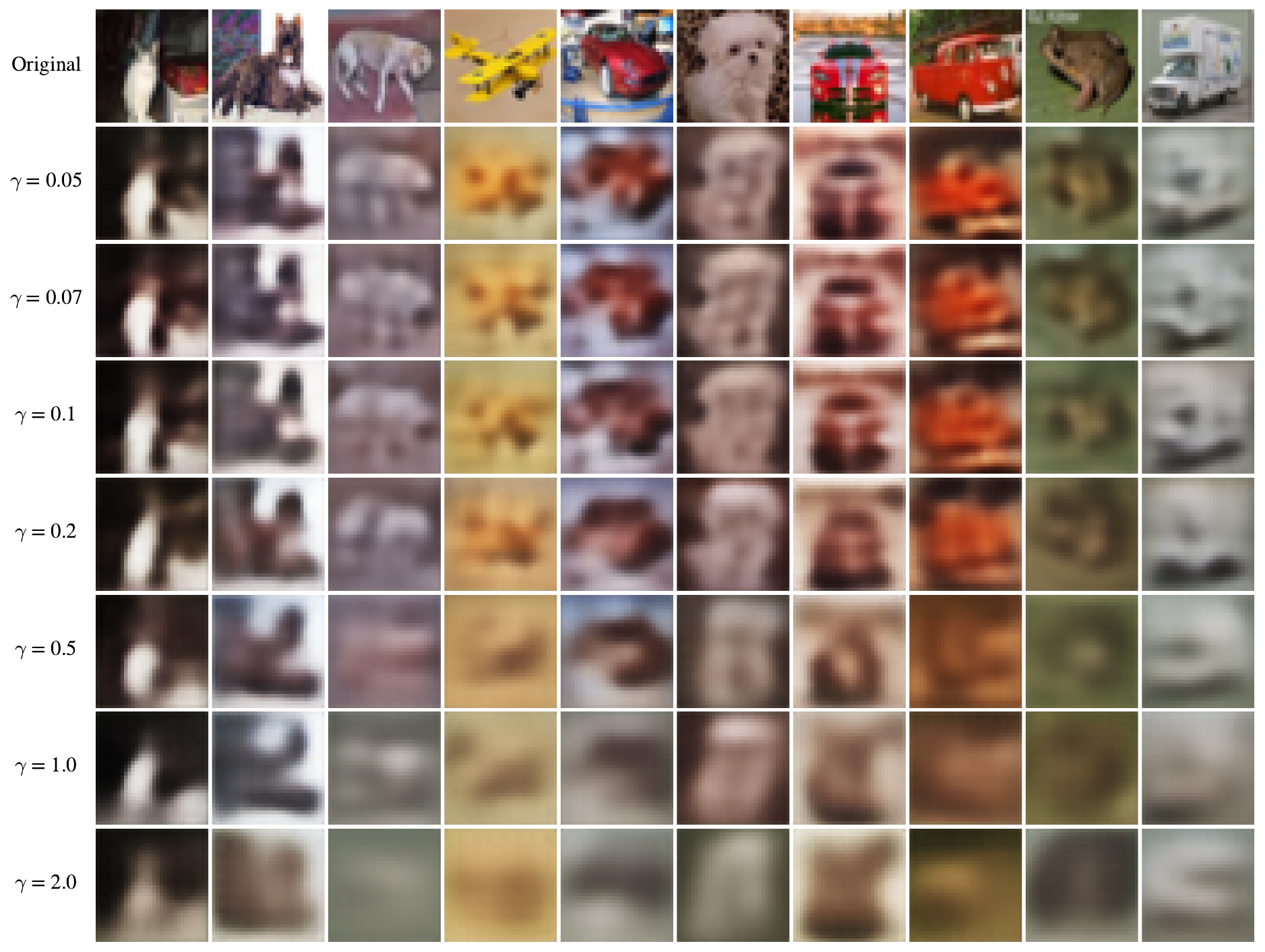}
        \caption{\label{fig:rd_sweep_recons_CIFAR-10_VAE} Reconstructions of CIFAR-10 samples with a VAE at different $\gamma$.}
    \end{subfigure}
    \caption{\label{fig:rd_sweep_VAE} Rate-distortion curves and reconstructions with a VAE for MNIST and CIFAR-10. Reconstructions were generated with $\DimZ=20$ for MNIST, and $\DimZ=128$ for CIFAR-10.}
\end{figure}

\begin{figure}
    \centering
    \begin{subfigure}[t]{1.0\linewidth}
        \centering
        \includegraphics[width=\linewidth]{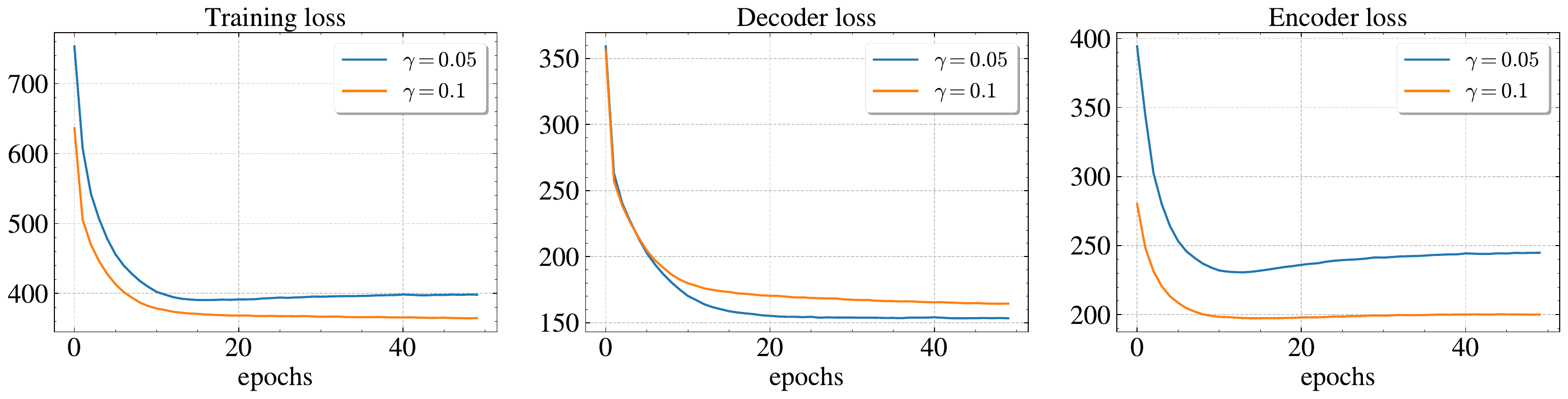}
        \caption{\label{fig:TIN_training_loss} \textbf{(Left)} Training loss of dAE on Tiny ImageNet, and its constituents, \textbf{(Middle)} the decoder loss (cf.\ \cref{eq:DecoderVAE}), and \textbf{(Right)} the encoder loss (cf.\ \cref{eq:LossEMDenoising}). The smooth, nearly monotone convergence of the training loss is to be expected from the similarities between \cref{alg:TrainingDiffEncAE} and expectation-maximization. See \cref{sec:EquilibratingToPosterior,sec:Experiments}.}
    \end{subfigure}
    \\[1em]
    \begin{subfigure}[t]{1.0\linewidth}
        \centering
        \includegraphics[width=\linewidth]{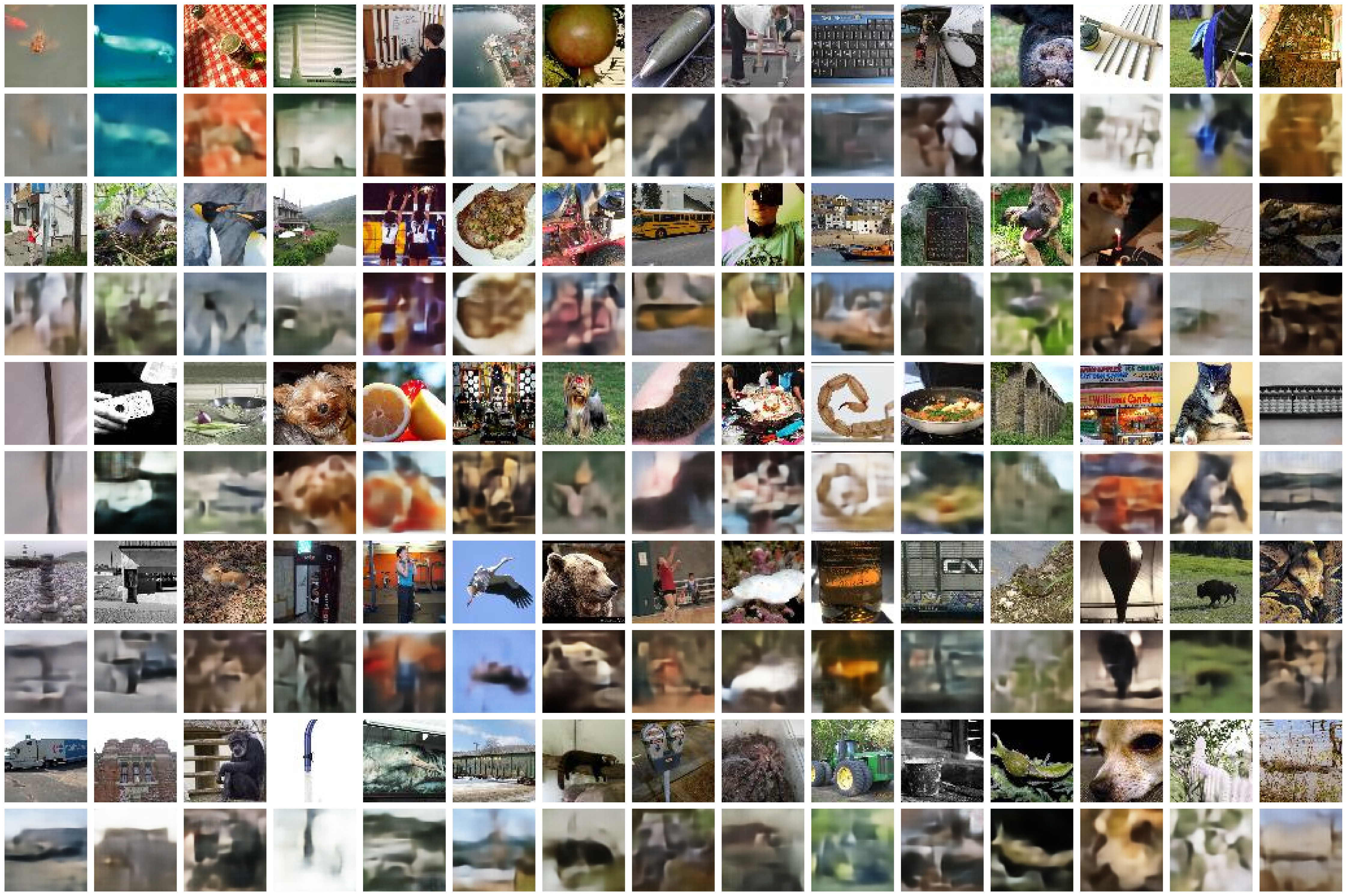}
        \caption{\label{fig:TIN-recons} Reconstructions of Tiny ImageNet with a dAE. Originals/reconstructions appear in odd/even rows.}
    \end{subfigure}
    \caption{\label{fig:TIN_loss_recons} Tiny ImageNet training and reconstructions with a diffusion encoder + convolutional decoder (dAE), and latent dimensions $D_Z = 512$.}
\end{figure}

\begin{figure}
    \centering
    \includegraphics[width=\linewidth]{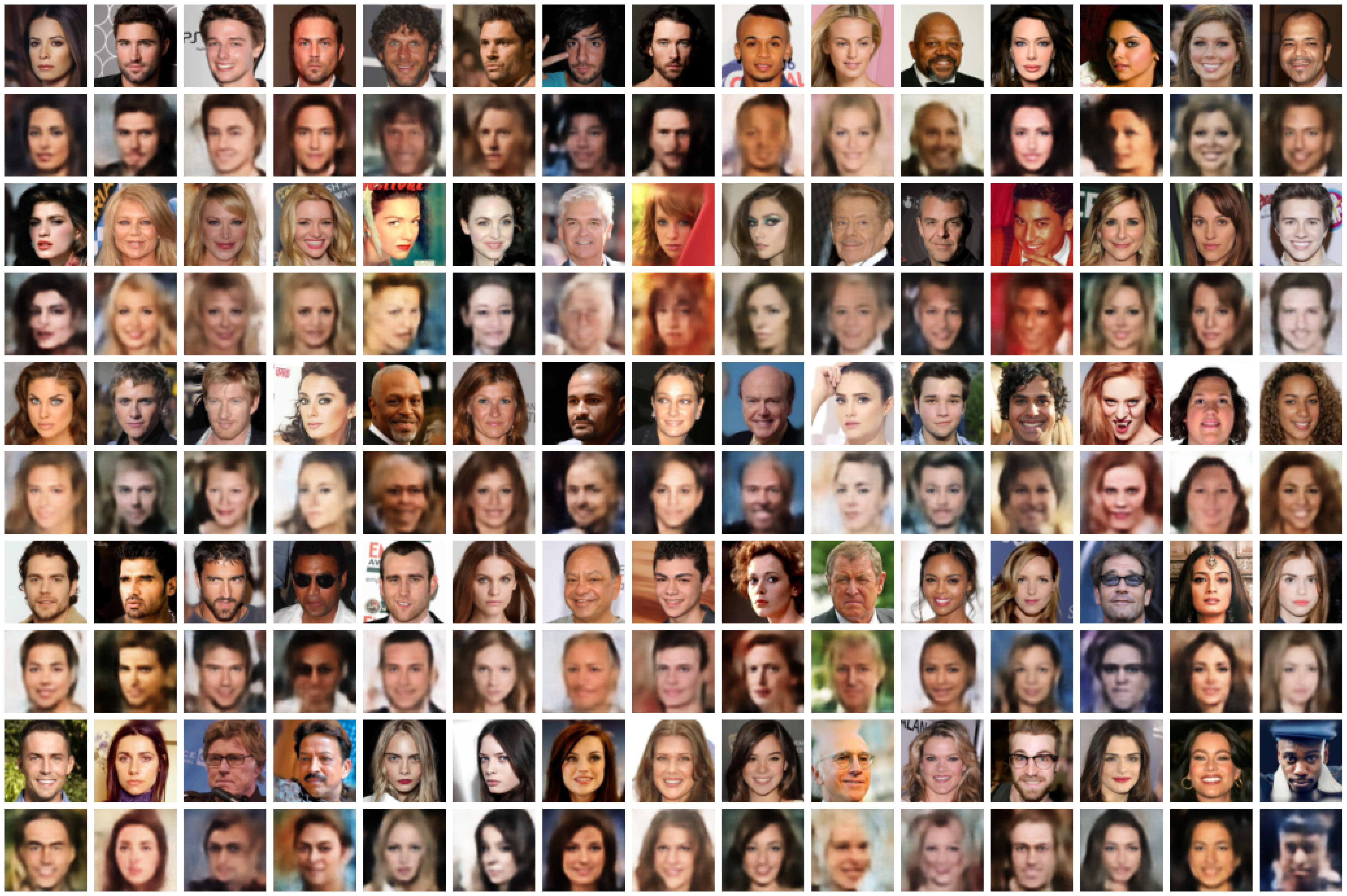}
    \caption{\label{fig:CelebA-recons} Reconstructions of CelebA-HQ (rescaled to $64 \times 64
    $) with a diffusion encoder + convolutional decoder (dAE). Originals/reconstructions appear in odd/even rows.}
\end{figure}

\begin{figure}
    \centering
    \includegraphics[width=\linewidth]{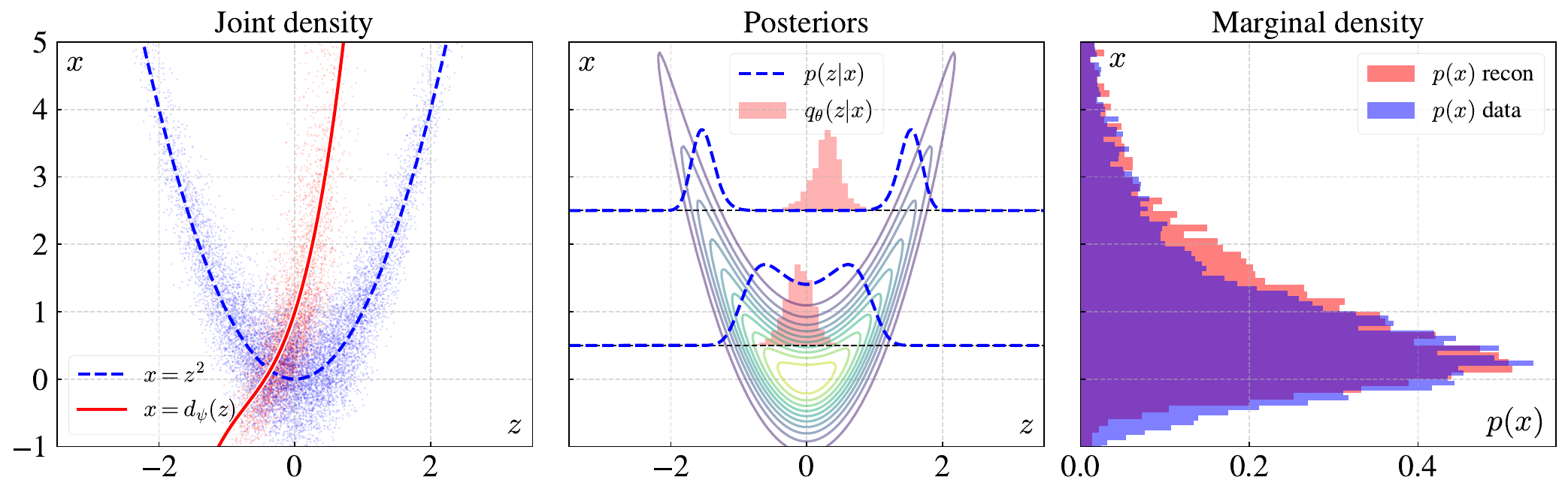}
    \caption{\label{fig:ToyModeldAE} A dAE applied to the toy model from \cref{sec:IBandVAEs}. \textbf{Left:} Samples from the true joint density $p(x,z)$ (blue) and the learned one (red). \textbf{Middle:} The true posterior at two values of $x$, and the diffusion encoder that best fits it. \textbf{Right:} The true marginal density and the one reconstructed by the dAE. The dAE admits more of the quadratic term $\psi_2 z^2$ from the polynomial decoder, which is why $d_\psi(z)$ is `more symmetric' under $z \to -z$ compared to \cref{fig:ToyModelVAE}. The learned posteriors are also nearly symmetric, just like $p(z|x)$, but they still have a roughly Gaussian silhouette. This is even though the diffusion encoder is \textit{fully capable of capturing $p(z|x)$}. See \cref{sec:Experiments} for discussion.}
\end{figure}

\newpage 
\section{Variational Autoencoders}
\label{sec:VAE}

We review the variational autoencoder (VAE) framework here, for completeness.

\subsection{ELBO}

We derive the training objective for a VAE, following \cite{Kingma19}. Given some data $\{ \vx \}_{i=1}^{N}$, we would like to model the underlying probability distribution that produced these samples. We assume each observation $\vx$ arises from an unobserved latent variable $\vz$ via the generative process $\vz \sim p(\vz)$, and $\vx \sim p_\ps(\vx|\vz)$. The marginal likelihood of $\vx$ under this model is obtained by integrating over all possible values of $\vz$,
\begin{equation*}
    \log p_\ps(\vx) = \log \int \d \vz \, p_\ps(\vx|\vz) p(\vz) = \log \int \d \vz \, p_\ps(\vx, \vz) .
\end{equation*}
However, this integral is computationally infeasible. We instead derive a tractable lower bound by introducing a variational distribution $q_\ph(\vz | \vx)$. Multiplying and dividing the integrand with this object, we get
\begin{equation*}
    \log p_\ps(\vx) 
    = \log \int \frac{q_\ph(\vz | \vx)}{q_\ph(\vz | \vx)}\, p_\ps(\vx, \vz)\, \d \vz
    = \log \E_{q_\ph(\vz | \vx)}\left[ \frac{p_\ps(\vx, \vz)}{q_\ph(\vz | \vx)} \right] ,
\end{equation*}
to which we can apply \textit{Jensen's inequality} and obtain
\begin{equation*}
    \log \E_{q_\ph(\vz | \vx)}\left[ \frac{p_\ps(\vx, \vz)}{q_\ph(\vz | \vx)} \right]
    \geq \E_{q_\ph(\vz | \vx)}\left[ \log \frac{p_\ps(\vx, \vz)}{q_\ph(\vz | \vx)} \right] .
\end{equation*}
This gives the Evidence Lower Bound (ELBO),
\begin{align}
    \log p_\ps(\vx)
        &\geq \E_{q_\ph(\vz | \vx)}\left[ \log p_\ps(\vx, \vz) - \log q_\ph(\vz | \vx) \right] \\
        &= \E_{q_\ph(\vz | \vx)}\left[ \log p_\ps(\vx | \vz) + \log p(\vz) - \log q_\ph(\vz | \vx) \right]
        \label{eq:IntermediateELBO} \\
        &= \underbrace{\E_{q_\ph(\vz | \vx)}[\log p_\ps(\vx | \vz)]}_{\text{Reconstruction term}}
        \; - \;
        \underbrace{\KL(q_\ph(\vz | \vx) \,\|\, p(\vz))}_{\text{Regularization term}}
        \label{eq:ELBO}
\end{align}
Maximum likelihood estimation of the parameters $\ph$ and $\ps$ follows from maximizing the r.h.s. Equivalently, we can minimize its negative,
\begin{equation}
    (\ph, \ps) =
    \operatorname{argmin}_{\ph, \ps}
    \left\{
        \E_{q_\ph(\vz | \vx)}[-\log p_\ps(\vx | \vz)]
        \; + \;
        \gamma  \KL(q_\ph(\vz | \vx) \,\|\, p(\vz))
    \right\} .
    \label{eq:LossVAE}
\end{equation}
Note that we have introduced a $\gamma > 0$ to adjust the regularization term \cite{Higgins17}; if $\gamma$ is large, the KL term forces all $q_\ph(\vz | \vx)$ to collapse to the same distribution $p(\vz)$, which is usually a Gaussian, and the VAE becomes unable to capture sufficient information from $\vx$ for reconstruction. A smaller $\gamma$ allows $q_\ph(\vz | \vx)$ to deviate farther from $p(\vz)$, trading off compression for reconstruction fidelity. Therefore, the parameter $\gamma$ plays the role of a temperature---higher $\gamma$ scrambles the encoder distribution, and lower values allow it to be more structured.

\subsection{Model components}

To implement the ELBO as a training loss, we choose the latent prior $p(\vz)$, the encoder $q_{\phi}(\vz|\vx)$, and the decoder $p_\ps(\vx | \vz)$ to be Gaussians.

\paragraph{Encoder} The encoder is a network that maps an input, say $\vx \in \sR^{\DimX}$, to $\mu_\ph(\vx) \in \sR^\DimZ$ and $\log \sigma_\ph(\vx)^2 \in \sR^\DimZ$, which parameterize a Gaussian
\begin{equation*}
    q_\ph(\vz | \vx) = \N(\vz; \mu_\ph(\vx), \operatorname{diag}(\sigma_\ph(\vx)^2)) .
\end{equation*}
The latent prior is $p(\vz) = \mathcal{N}(0, I)$. The KL divergence between two diagonal Gaussians, $p_1(\vx) = \mathcal{N}(\vx; \mu_1, \Sigma_1)$ and $p_2(\vx) = \mathcal{N}(\vx; \mu_2, \Sigma_2)$, has a closed-form expression
\begin{equation*}
    \KL(p_1 \| p_2) = \frac{1}{2}\left[\log\frac{|\Sigma_2|}{|\Sigma_1|} - d + \text{tr} \{ \Sigma_2^{-1}\Sigma_1 \} + (\mu_2 - \mu_1)^\top \Sigma_2^{-1}(\mu_2 - \mu_1)\right] .
\end{equation*}
With $\mu_1 = \mu_\ph$, $\mu_2 = 0$, $\Sigma_1 = \operatorname{diag}(\sigma_\ph^2)$, and $\Sigma_2 = I$, we get
\begin{align*}
    \text{tr} \{ \Sigma_2^{-1}\Sigma_1 \} &= \sum_{j=1}^{d} \sigma_j^2 \\
    (\mu_2 - \mu_1)^\top \Sigma_2^{-1}(\mu_2 - \mu_1) &= \mu_\ph^\top I \mu_\ph = \sum_{j=1}^{d} \mu_j^2 \\
    \log\frac{|\Sigma_2|}{|\Sigma_1|} &= -\log \prod_{i=1}^{D} \sigma_j^2 .
\end{align*}
Thus, the regularization term in the ELBO is
\begin{equation}
    \KL (q_\ph(\vz | \vx) \,\|\, p(\vz)) =
        \frac{1}{2} \sum_{j=1}^{\DimZ} ((\mu_\ph^j)^2 + (\sigma_\ph^j)^2 - \log (\sigma_\ph^j)^2 - 1) ,
\end{equation}
where $\mu_\ph^j$ and $\sigma_\ph^j$ are the $j^{\rm th}$ components of $\mu_\ph$ and $\sigma_\ph$. In practice, the encoder outputs log-variance $\log \sigma_\ph(\vx)^2$, which we exponentiate to get $\sigma_\ph(\vx)^2$.

\paragraph{Decoder} Assume the decoder defines a Gaussian distribution with fixed variance:
\begin{equation*}
    p_\ps(\vx | \vz) = \N(\vx; d_\ps(\vz), \sigma^2 I) \label{eq:GaussianLikelihood}
\end{equation*}
Taking the log of this likelihood:
\begin{equation}
    \log p_\ps(\vx | \vz) = -\frac{1}{2\sigma^2} \|\vx - d_\ps(\vz)\|^2 + \text{const}
\end{equation}
Since the constant and $\sigma^2$ can be absorbed into optimization, the reconstruction term becomes:
\begin{equation}
    \E_{q_\ph(\vz | \vx)}[-\log p_\ps(\vx | \vz)]
        \approx \E_{\vz \sim q_\ph(\vz | \vx)}[ \|\vx - d_\ps(\vz)\|^2 ] .
    \label{eq:DecoderVAE}
\end{equation}
So the reconstruction loss is mean squared error (MSE). In each training cycle the $\vz$ values are generated from the training data $\vx$ using $\vz = \mu_\ph(\vx) + \sigma_\ph(\vx) \odot \bm{\epsilon}$ where $\bm{\epsilon} \sim \N(0, I)$. This is called the \textit{reparameterization trick}, and it allows the gradients to flow through the sampling step, since each $\vz$ is a function of $\ph$.

\section{Adjoint Matching}
\label{sec:AdjointMatching}

We briefly review the \textit{adjoint matching} method, introduced in \cite{Domingo25}, which improves on classic adjoint methods for training neural SDEs. We begin with a base generative process
\begin{equation}
    \d \mZ_\t = f_{\rm b}(\mZ_\t, \t) \d \t + \g(\t) \d \mB_\t , \qquad \mZ_0 \sim q_0 = \N(0, I) .
\end{equation}
The goal is to fine-tune it by adding a learned control $\u_\th(z, t)$ such that the controlled process,
\begin{equation}
    \d \mZ_\t = \left[ f_{\rm b}(\mZ_\t, \t) + \g(\t) \u_\th(\mZ_\t, \t) \right] \d \t + \g(\t) \d \mB_\t ,
    \label{eq:ControlledSDEAM}
\end{equation}
generates samples from a \textit{tilted} distribution at time $\T$,
\begin{equation}
    q(\vz) \propto q_{\rm base}(\vz) e^{r(\vz)} \label{eq:Tilted}
\end{equation}
where $r$ is some reward. This is accomplished by minimizing the adjoint matching objective
\begin{equation}
    \begin{gathered}
        \L_{\rm AM}(\u_\th; \pZ)
            := \frac{1}{2} \int_{0}^{\T} \d \t \, \| \u_\th(\vz_\t, \t) + \g(\t) \ta(\t; \pZ) \|^2,
        \quad \pZ \sim q_{\bar{\u}}, \ \bar{\u} = \texttt{stopgrad}(\u_\th) , \\
        \text{where} \quad
        \frac{\d}{\d \t} \ta(\t; \pZ) = - \ta(\t; \pZ)^\top \nabla f_{\rm b}(\mZ_\t, \t) , \\
        \ta_\T = - \nabla r(\mZ_\T) .
    \end{gathered}
    \label{eq:AdjointMatching}
\end{equation}
Here $\pZ$ denotes an entire path $\pZ = (\mZ_\t)_{t \in [0,\T]}$ generated by the controlled process \cref{eq:ControlledSDEAM}, with $\bar{\u}$. $\L_{\rm AM}(\u_\th; \mZ)$ is optimized with a gradient-based method, but $\mZ_\t$ and $\ta_\t$ should be computed without gradients. Two key features in \cref{eq:AdjointMatching} are worth highlighting. First, the functional landscape of $\L_{\rm AM}$ is convex, which leads to more stable training compared to the highly non-convex standard adjoint objective \cite{Li20}. Second, the adjoint ODE does \textit{not} involve $\nabla \u_\th$, unlike the standard adjoint method. This makes the ODE cheaper to compute, which is why $\ta$ is called the \textit{lean adjoint}.

The relevance of this method becomes apparent when we note that the optimal encoder from \cref{eq:OptimalEncoderVAE} has precisely the same form as \cref{eq:Tilted}. Indeed, setting $f_{\rm b} = f_{\rm eq}$, $f_\th = f_{\rm eq} + \g \u_\th$, and $r = \frac{1}{\gamma} \log \P_\ps(\vx|\vz_\T)$ (cf.\ \cref{eq:PathKL}),
\begin{equation}
    \min \E \left[ \KL(\mathbb{Q} \| \mathbb{P}) - \frac{1}{\gamma} \log p_\ps(\vx|\vz_\T) \right]
    \rightarrow
    \min \E \left[ \frac{1}{2} \int_{0}^{\T} \d t \|\u_\th(\vz_\t, \t; \vx)\|^2 - r(\vz_\T; \vx) \right] .
\end{equation}
The latter is precisely the starting point from which \cref{eq:AdjointMatching} is derived in \cite{Domingo24}. Note that we have introduced $\vx$-dependence into the control in accordance with the discussion in \cref{sec:StochasticEncoder}. Once again, we see an encoder training objective that includes regularization pressure, just like \cref{eq:DEncRegularizationSingle}. For this reason, adjoint-matching is poised to suffer from synchronization issues similar to those described in \cref{sec:RegularizationOnEncoderSide}---the encoder and decoder can fall out of alignment since they are pulling the latent in opposing directions in each decoupled update.

Both adjoint matching and our diffusion encoder require the target $\mZ_\T$ to be frozen. A key difference, compared to our diffusion encoder, is that we must store the entirety of each path $\pZ$ to integrate the adjoint ODE, and optimize $\L_{\rm AM}$. In practice, the paths are generated by an Euler-Maruyama solver (cf.\ \cref{eq:SolutionEM}), with a finite number of steps. On the other hand, the diffusion loss is trained with forward noised samples rather than the adjoint, and these can be produced without solving the forward SDE (cf.\ \cref{eq:PropagatedVP}); this is simulation-free training. We also do not need to store the trajectories from the encoding step.

\section{Encoder Architecture}
\label{sec:EncoderArchitecture}

At the heart of our diffusion encoder is a neural network $\e_\th(\tz_\s, \s; \vx) \equiv \e_\th(\vz_\t, \T-\t; \vx)$. It accepts three inputs: the noised latent $\tz_\s \in \sR^{\DimZ}$, a conditioning image $\vx \in \mathbb{R}^{\etH \times \etW \times \etC}$, and the forward time $\s \in (0, \T]$. Its output is an estimate of the entropy-matching vector, and has the same shape as $\tz_\s$.

The network $\e_\th$ controls the evolution of the initial distribution $q(\vz_0)$ to the posterior $q(\vz_\T | \vx)$ (cf.\ \cref{eq:EntropyMatchingSDE}). The value of $\e_\th$ at each time slice parameterizes an $\vx$-dependent Gaussian kernel (see \cref{sec:StochasticEncoder}). Thus, the network effects a continuum of minuscule transformations that gradually encode information from $\vx$ into $\vz_\T$. The encoder architecture must, therefore, include an effective mechanism to compress $\vx$ into $\vz_\t$ in a $\t$-dependent way. This can be accomplished with attention \cite{Vaswani17}.

\subsection{Attention}

To review, cross-attention is the transformation
\begin{equation}
    \rmY(\rmZ) = \Attn(\rmQ, \rmK, \rmV)
        \equiv \softmax \left[ \frac{\rmQ \rmK^\top}{\sqrt{\DimKey}} \right] \rmV ,
    \label{eq:Attention}
\end{equation}
where $\rmZ \in \sR^{N_Q \times \DimZ}$ and $\rmX \in \sR^{N_K \times \DimX}$, with
\begin{equation}
    \begin{aligned}
        \rmQ &= \rmZ \rmW_Q \in \sR^{N_Q \times \DimKey} , \quad \rmW_Q \in \sR^{\DimZ \times \DimKey} \\
        \rmK &= \rmX \rmW_K \in \sR^{N_K \times \DimKey} , \quad \rmW_K \in \sR^{\DimX \times \DimKey} \\
        \rmV &= \rmX \rmW_V \in \sR^{N_K \times \DimVal} , \quad \rmW_V \in \sR^{\DimX \times \DimVal}
    \end{aligned}
    \label{eq:CAttnDimensions}
\end{equation}
Therefore, $\rmQ \rmK^\top$ has dimension $N_Q \times N_K$, and the softmax is taken over the key (column) axis. Each row can be viewed as the conditional probability of the keys, given the query. Each of the $N_Q$ queries is assigned $N_K$ probabilities this way. These probabilities are then used to linearly combine the $N_K$ value vectors in $\rmV$, each of $\DimVal$ dimensions, to produce the final output of shape $N_Q \times \DimVal$. The weights $\rmW_\bullet$ are learnable parameters. In self-attention $\rmZ = \rmX$.

In multi-head attention, there are $H$ attention heads, which are combined as
%
\begin{equation}
    \rmY(\rmZ) = \textrm{Concat} [\rmH_1, \dots, \rmH_H] \rmW_O , \quad \textrm{where} \
    \rmH_h = \Attn(\rmQ_h, \rmK_h, \rmV_h) .
    \label{eq:MultiheadAttention}
\end{equation}
Each $\rmH_h$ has dimension $N_Q \times \DimVal$, so the concatenated matrix has dimension $N_Q \times H \DimVal$. This is transformed by the linear matrix $\rmW_O$ of dimension $H \DimVal \times \DimZ$ to produce $\rmY$ of dimension $N_Q \times \DimZ$, which is the same as that of the original input matrix $\rmZ$. Each position in $\rmX$ looks at all positions in $\rmZ$, computes relevant probabilities, and pulls out a weighted combination of information from $\rmZ$. In short, ``$\rmZ$ attends to $\rmX$.''

For our purposes, the key point is that attention is an input-dependent transformation. For example, by using $\vz_\t$ to attend to $\vx$, we can extract different levels of information about $\vx$ depending on where we are along the transformation $\vz_0 \to \vz_\T$. To make this precise, consider the problem of encoding an image of shape $[\etH, \etW, \etC]$ into a 1D feature vector with `\feat' number of components. First, we flatten the image so that $[\etH, \etW, \etC] \to [\etH \etW, \etC]$. This will be $\rmX$, so $N_K = \etH \etW, \DimX = \etC$. Next, recall that the final output $\rmY(\rmZ)$ has the same shape as $\rmZ$, so we choose the query $\rmZ$ to have the same dimensions as the desired feature vector. Therefore, $N_Q = 1, \DimZ = \feat$. If we choose $H = 4$ heads in \cref{eq:MultiheadAttention}, then $4 \DimVal = \DimZ \implies \DimVal = \feat/4$. With these choices, $\rmZ \in \R^{1 \times \feat}$ and $\rmX \in \R^{\etH \etW \times \etC}$, and \cref{eq:CAttnDimensions} is
\begin{equation}
    \begin{aligned}
        \rmQ &= \rmX \rmW_Q \in \sR^{1 \times \DimKey} , \quad \rmW_Q \in \sR^{\feat \times \DimKey} \\
        \rmK &= \rmZ \rmW_K \in \sR^{\etH \etW \times \DimKey} , \quad \rmW_K \in \sR^{\etC \times \DimKey} \\
        \rmV &= \rmZ \rmW_V \in \sR^{\etH \etW \times \feat/4} , \quad \rmW_V \in \sR^{\etC \times \feat/4}
    \end{aligned}
    \label{eq:CAttnDimensionsExample}
\end{equation}
We will choose $\DimKey$  shortly. For now, notice that $\rmQ \rmK^\top \in \R^{1 \times \etH \etW}$ is a row vector. Then, $\softmax[\rmQ \rmK^\top / \sqrt{\DimVal}]$ can be interpreted as probabilities over $\etH \etW$ pixels---it weighs how much each pixel influences ($1/4$ of) the final feature vector. The rows of $\rmV$ are $\etH \etW$ `basis' vectors built up from the image $\rmX$ itself---each of the $\etH \etW$ pixels in the image is converted to a basis vector of length $\feat/4$ by $\rmW_V$, and then mixed according to the $\etH \etW$ probabilities from the softmax.

Having multiple attention heads allows the same pixels to be mapped to different feature basis vectors. These basis vectors can also get mixed in different proportions since the softmax in a different head can produce a new set of probabilities for the same $\rmX$ and $\rmZ$. So each $\feat/4$ segment in $\rmY$ can encode different information about the image.
Finally, we turn to the $\rmW_Q$ and $\rmW_K$ matrices. In \cref{eq:Attention}, $\rmW_Q \rmW_K^\top$ induces a rank-$\DimKey$ attention bottleneck on the logits. A similar bottleneck also exists in the value pathway, $\rmW^{h}_V \rmW^{h}_O$, where $\rmX$ is projected down to $\DimVal$ and then back up to $\DimZ$.

The calculation above is meant as an illustration. In practice, it is wasteful to attend to the input image $\vx$ at its native resolution. Instead, we run the image through a shallow convolutional network to scale it from $\etH \times \etW \times \etC \to (\etH/4) \times (\etW/4) \times \etC'$ (see \cref{sec:WeakImageEncoder}). The query vector can be $\vz_\t$ or an embedding of the time variable itself.

\subsection{Input Embeddings}
\label{sec:InputEmbeddings}

The time variable is lifted into a higher-dimensional feature space using random Fourier features, which are fixed at initialization and receive no gradient updates. This is particularly beneficial for low-dimensional inputs, which would otherwise be spectrally underrepresented in the network's feature space \cite{Tancik20}. Concretely, for a signal $u$ and a frozen random matrix $\rmB$ drawn from $\mathcal{N}(0, I)$ and scaled by $\gamma$, the embedding is
\begin{equation}
    \phi(u; \rmB) = \bigl[\sin(2\pi u \rmB^\top),\, \cos(2\pi u \rmB^\top)\bigr]
    \in \sR^{2M},
    \label{eq:fourier}
\end{equation}
where $M$ is the \textit{mapping size}. For the time coordinate $\s$, we use a dedicated matrix $\rmB_\s \in \sR^{M \times 1}$ with scale $\gamma_\s = 10$, giving a $2M$-dimensional time embedding. Optionally, in cases where $\DimZ$ is low, we also found it useful to embed the latent $\vz_\t$, with its own $\rmB_\vz \in \sR^{M \times \DimZ}$ with scale $\gamma_\vz = 10$. Both embeddings are then projected to width $M$ via a learned linear layer followed by a SiLU activation.

\subsection{Weak Image Encoder}
\label{sec:WeakImageEncoder}

The input image $\vx$ is downsized with a shallow convolutional feature extractor, which we call the
\emph{weak encoder}, before it is sent to the attention block. This stage serves to reduce the computational overhead of attending to the whole image $\vx$, without compressing away too much information from it. It consists of two strided convolutional layers, each with a $4 \times 4$ kernel, stride 2, and \textsc{same} padding, activated by SiLU:
\begin{equation}
    \vx_{\rm enc}
        = \mathrm{WeakEncoder}(\vx)
        \in \mathbb{R}^{\etH/4 \times \etW/4 \times \etC'} .
\end{equation}
%
The spatial grid is then flattened to produce a sequence of $N_K = \etH/4 \times \etW/4$ `patch' tokens of dimension $\etC'$,
\begin{equation}
    \vx_{\rm flat}
        = \mathrm{reshape}(\vx_{\rm enc})
        \in \mathbb{R}^{N_K \times C'}.
    \label{eq:xFlat}
\end{equation}
This is the input $\rmX$ to the attention block in \cref{eq:Attention}. By scaling down the spatial dimensions by a factor of $4$ in each direction, we reduce the sequence length $N_K \to N_K/16$, shrinking the cost of the cross-attention operation by the same factor since it is linear in $N_K$ (cf.\ \cref{eq:CAttnDimensionsExample}).

\paragraph{Positional encodings and multi-scale tokens} The raw sequence in \cref{eq:xFlat} is permutation-invariant---patches at opposite corners of the image are distinguishable only by content, not by location. We therefore inject 2-D sinusoidal positional encodings into the fine-scale grid $\etH/4 \times \etW/4$ and concatenate a coarse-scale token sequence obtained by average pooling to $\etH/8 \times \etW/8$ before the same positional encoding step. Both token sequences are passed through a shared linear projection before being concatenated,
\begin{equation}
    \vx_{\rm tokens}
    = \bigl[ \vx_{\rm fine} + \mathrm{PE_{fine}}\,;\;
             \vx_{\rm coarse} + \mathrm{PE_{coarse}} \bigr]
    \in \sR^{(N_K + N_K/4) \times D_{\rm tok}},
\end{equation}
where $N_K = (\etH/4)^2$ and $N_K/4 = (\etH/8)^2$ for square inputs. The dual-scale design provides both fine-grained texture detail and low-frequency layout information at a 25\% overhead in token count.

\subsection{Cross-Attention Image Conditioning}
\label{app:architecture:attention}

The projected time embedding $\rve_\s \in \mathbb{R}^M$ from \cref{sec:InputEmbeddings} serves as the single query vector for multi-head cross-attention against the image token sequence:
\begin{equation}
    \vx_{\rm attended}
    = \mathrm{MultiHeadAttn}\!\bigl(\rmQ = \rve_\s[{:},\,\mathrm{None},{:}],\;
                                    \rmK = \vx_{\rm flat},\;
                                    \rmV = \vx_{\rm flat}\bigr)
    \in \mathbb{R}^{M},
\end{equation}
with $H = 4$ heads and key/value dimension $M$. Setting the key--value dimension equal to $M$ ensures no information bottleneck is introduced through the projection matrices $\rmW_K$ and $\rmW_V$. The attended vector summarizes the patches that the current time state $\rve_\s$ finds relevant, conditioning the subsequent score network on that selection. The global conditioning vector is then assembled from the time embedding and the attended image summary
\begin{equation}
    \vc = \mathrm{Dense}\bigl(\mathrm{SiLU}( [\rve_\s;\, \vx_{\rm attended}]) \bigr)
    \in \mathbb{R}^M,
    \label{eq:TimeAndAttended}
\end{equation}
where $[\,\cdot\,;\,\cdot\,]$ denotes concatenation across the feature axis. In the simplest design, this vector $\vc$ is computed once and reused at every residual block in the backbone described below. A more expressive alternative---performing cross-attention at every block using the current hidden state as the query---is described at the end of the next section.

\subsection{MLP Backbone with Adaptive Conditioning}
\label{sec:MLPBackbone}

The backbone is a depth-$L$ residual MLP with hidden widths given by the sequence $\texttt{features} = [d_1, \ldots, d_L]$. Each block $i = 1, \ldots, L$ applies the following operations to the current hidden state $\vh$:
\begin{enumerate}
    \item \textbf{Projection.} $\vh \leftarrow \mathrm{Dense}(d_i)(\vh)$.
    \item \textbf{Layer normalization.} $\vh \leftarrow \mathrm{LayerNorm}(\vh)$, standardizing each sample to zero mean and unit variance.
    \item \textbf{Adaptive modulation (FiLM/AdaLN).} The conditioning vector $\vc^{(i)}$ is projected to a scale--shift pair $(\rho_i, \beta_i) \in \sR^{d_i} \times \mathbb{R}^{d_i}$ via a single linear layer, and applied as
    \begin{equation}
      \vh \leftarrow \vh \odot (1 + \rho_i) + \beta_i.
      \label{eq:film}
    \end{equation}
    Multiplying by $(1 + \rho_i)$ rather than $\rho_i$ alone initializes the modulation near the identity, which improves the stability of the training. This is the Adaptive Layer Normalization (AdaLN) design used in DiT \cite{Peebles22}.
    \item \textbf{Activation.} $\vh \leftarrow \mathrm{SiLU}(\vh)$.
    \item \textbf{Residual connection.} If $\vh_{\rm in}$ and $\vh$ share the same width, $\vh \leftarrow \vh + \vh_{\rm in}$; otherwise, a learned linear projection aligns the dimensions before addition.
\end{enumerate}
After the final block, a linear read-out projects $\vh$ to $\sR^{\DimZ}$, producing the score network output $\e_\th$.

\paragraph{Per-block cross-attention} When $\vc$ is fixed across all blocks as in \cref{eq:TimeAndAttended}, the image context is computed from the initial latent embedding and reused at every depth. As the hidden state $\vh$ evolves through successive nonlinear transformations, this fixed summary grows increasingly misaligned with the current representation. To address this, the improved architecture performs cross-attention at every residual
block, using the current hidden state as the query:
\begin{equation}
    \vx_{\rm attended}^{(i)}
    = \mathrm{MultiHeadAttn}^{(i)}\!\bigl(\vh^{(i)},\; \vx_{\rm tokens}\bigr),
    \qquad
    \vc^{(i)} = \mathrm{Dense}^{(i)}\!\bigl([\rve_\s;\, \vx_{\rm attended}^{(i)}]\bigr),
\end{equation}
so that the scale and shift parameters of every FiLM modulation are derived from an image context that is semantically coherent with $\vh^{(i)}$. Each block maintains its own Q/K/V/O projection weights.

\subsection{Decoder Architecture}
\label{sec:DecoderArchitecture}

The decoder maps a latent vector $\vz$ back to image space through a sequence of convolutional upsampling stages. The latent is first projected by a dense layer to a small spatial grid, which is then progressively upsampled using transposed convolutions, each followed by a residual refinement block. Optionally, self-attention blocks with layer normalization can be inserted at low-resolution feature maps to capture long-range spatial dependencies during upsampling. A final transposed convolutional layer maps the feature map to the output image dimensions, with a sigmoid activation to constrain pixel values to $[0,1]$.

\subsection{Hyperparameter Summary}
\label{app:architecture:hyperparams}

\cref{tab:ArchHyperparameters} collects the architectural hyperparameters used in the experiments reported in~\cref{sec:Experiments}.

\begin{table}[h]
\centering
\caption{Encoder architecture/training hyperparameters.}
\label{tab:ArchHyperparameters}
\begin{tabular}{lccc}
\toprule
Hyperparameter & MNIST & CIFAR-10 & Tiny ImageNet/CelebA \\
\midrule
Latent dimension $\DimZ$              & 20       & 128                      & 512                     \\
Mapping size $M$                    & 128      & 256                      & 512                     \\
MLP widths (\texttt{features})        & [512, 256] & [512, 512, 256, 256]   & [512, 512, 512, 256, 256] \\
Weak encoder channels               & [32, 64] & [64, 128]               & [128, 256]              \\
Attention heads $H$       & 4        & 8                        & 8                       \\
Fourier scale $\gamma_\vz$            & 10       & 10                       & 10                      \\
Fourier scale $\gamma_\s$            & 10       & 10                       & 10                      \\
$\vz$ Fourier embedding ($\vz_{\rm embed}$) & False & False                  & False                   \\
$\beta_{\min}$                      & 0.1      & 0.1                      & 0.1                     \\
$\beta_{\max}$                      & 16       & 12                       & 12                      \\
Training steps per $\vz^\star$ ($n_{\rm step}$) & 10 & 10                   & 10                      \\
Lantents per image ($B_\vz$)           & 4        & 4                        & 4                       \\
Encoder learning rate               & 0.001    & 0.001                    & 0.0002                  \\
Batch size                          & 256      & 128                      & 128                     \\
Epochs                              & 50      & 50                      & 50                     \\
\midrule
Time to train (1x H200)              & $\approx 15$ mins  & $\approx 45$ mins & $\approx 325$ mins                     \\
\bottomrule
\end{tabular}
\end{table}

\section{Notation}
\label{sec:Notation}

The natural logarithm is denoted by $\log$. Scalars are written in plain letters, while boldface symbols such as $\mX, \mY, \mZ$ denote higher-dimensional random variables. Lowercase $\vx \in \sR^{\DimX}$ is a realization of $\mX$. We use $\vz \in \sR^{\DimZ}$ to denote the latents, and $\vx$ to denote the inputs to the encoder. For image inputs, $\DimX$ is replaced with $\etH \times \etW \times \etC$.

\begin{figure}[ht]
    \pgfmathdeclarefunction{gauss}{2}{%
      \pgfmathparse{1/(#2*sqrt(2*pi))*exp(-((x-#1)^2)/(2*#2^2))}%
    }
    
    \centering
    \begin{tikzpicture}
    \begin{scope}[shift={(-5,0)}]
        \begin{axis}[grid=none,
            ymax=2.1,
              axis lines=middle,
              y=1cm,
            y axis line style={draw=none},
            x axis line style={{stealth'}-{stealth'}},
            ytick=\empty,
            xtick=\empty,
            enlargelimits,
            rotate=90,
            ]
            \addplot[violet,fill=violet!20,domain=-2.4:2.4,samples=200]  {2.5*gauss(0,0.85)} \closedcycle;
    
            \node[align=center] at (axis cs:2.2,1.5) {$q(\vz_0)$ \\ or \\ $q(\tz_\T)$};
            \node at (axis cs:-2.5,0.25) {$\vz$};
        \end{axis}
    \end{scope}
    
    \draw[thick,-{stealth'}] (4.4,5) -- ++(-6.4,0) coordinate[pos=0.5] (a);
    \node[align=center,shift={(0,1)}] at (a) {Forward SDE \\ $ \scriptstyle \d \tilde{\mZ}_{\s} = f_{\rm eq} \d \s + \g \d \hat{\B}_{\s}$};
    
    \draw[thick,-{stealth'}] (-2,1.75) -- ++(6.4,0) coordinate[pos=0.5] (b);
    \node[align=center,shift={(0,-1)}] at (b) {Reverse SDE \\ $ \scriptstyle \d \mZ_{\t} = -(f_{\rm eq} - \g^2 \nabla \log q) \d \t + \g \d \B_{\t}$};

    
    \begin{scope}[shift={(5,0)}]
        \begin{axis}[grid=none,
            ymax=2.1,
            axis lines=middle,
              y=1cm,
            y axis line style={draw=none},
            x axis line style={{stealth'}-{stealth'}},
            ytick=\empty,
            xtick=\empty,
            enlargelimits,
            rotate=-90,
            ]
            \addplot[violet,fill=violet!20,domain=-2.4:2.4,samples=200] {gauss(-1.1,0.2) + gauss(1.1,0.2)} \closedcycle;
    
            \node[align=center] at (axis cs:-2.2,1.5) {$q(\vz_\T|\vx)$ \\ or \\ $q(\tz_0|\vx)$};
            \node at (axis cs:2.5,0.25) {$\vz$};
        \end{axis}
    \end{scope}

    \begin{scope}[shift={(0,-1.5)}]
        \draw[-{stealth'}] (6.6,0) -- (-3.9,0) node[left] {$\s$};
        \draw (-2.69,0.25) node[above] {$T$} -- ++(0,-0.5);
        \draw (5.21,0.25) node[above] {$0$} -- ++(0,-0.5);
        \draw[|->] (5.21,-0.6) -- (0.03,-0.6) node[circle,fill=white,draw=none,pos=0.5] {$\s$};
    \end{scope}

    \draw[dashed] (0,-1) -- (0,-4);

    \begin{scope}[shift={(0,-3.5)}]
        \draw[-{stealth'}] (-3.9,0) -- (6.6,0) node[right] {$\t$};
        \draw (-2.69,0.25) -- ++(0,-0.5) node[below] {$0$};
        \draw (5.21,0.25) -- ++(0,-0.5) node[below] {$T$};
        \draw[|->] (-2.69,0.6) -- (-0.03,0.6) node[circle,fill=white,draw=none,pos=0.5] {$\t$};
    \end{scope}
    
    
    \end{tikzpicture}
    \caption{\label{fig:ReverseDiffusion}A schematic of the forward and reverse diffusion processes.}
\end{figure}
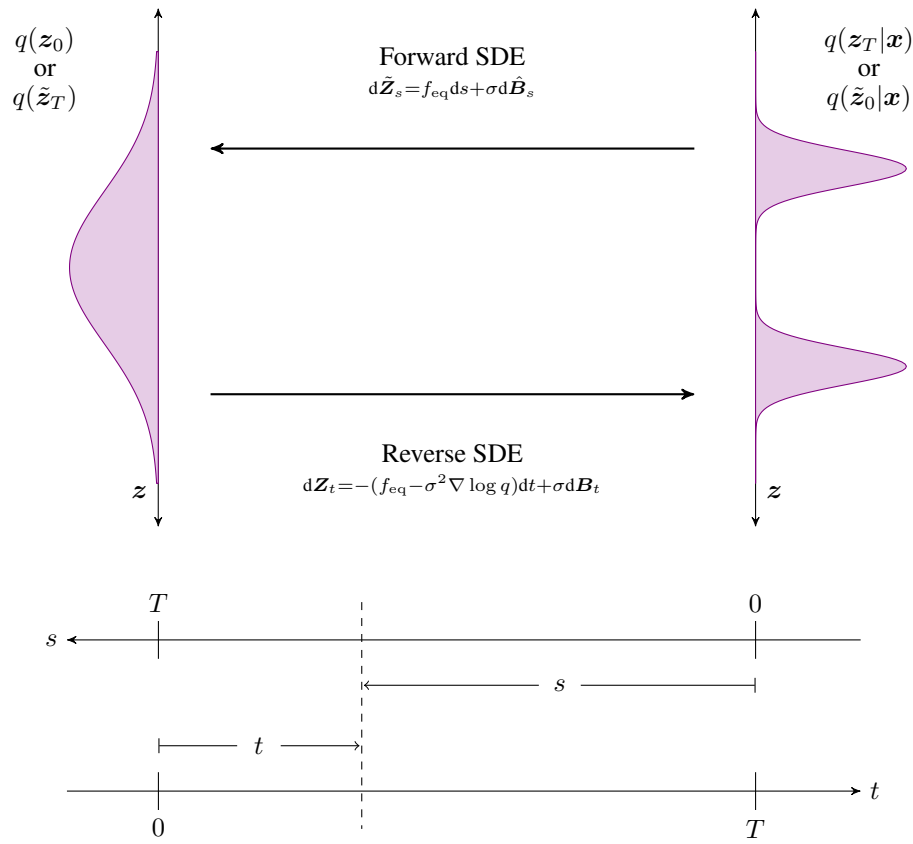

We use the time variable $\s$ for the forward diffusion process, which runs from left ($\s=0$) to right ($\s=\T$) in \cref{fig:ReverseDiffusion}.
$\hat{\B}_\s$ and $\B_\t$ denote the Brownian motions associated with the forward and reverse/controlled SDEs, respectively. $\nabla$ is the gradient with respect the latent coordinates, and $\pd_\t, \pd_\s$ are partial time derivatives.

The density $\P(\tz_\s, \s)$ is the same as $\P(\vz_\t, \t)$. That is, the symbol $\P$ is overloaded so we do not have to write $\P(\cdot, \s)=\P(\cdot, \T-\t)$ everywhere. Diffusion takes an infinite time to equilibrate, but we always take $\T$ to be large compared to the intrinsic time scale of the diffusion process. We use the symbol $S \deq -\int \P \log \P$ for differential entropy.

\newpage

\end{document}

%% file: math_commands.tex
\usepackage{amsmath,amsfonts,bm}

\def\eqref#1{equation~\ref{#1}}

\def\1{\bm{1}}

\def\rve{{\mathbf{e}}}

\def\rmB{{\mathbf{B}}}

\def\rmH{{\mathbf{H}}}

\def\rmK{{\mathbf{K}}}

\def\rmQ{{\mathbf{Q}}}

\def\rmV{{\mathbf{V}}}
\def\rmW{{\mathbf{W}}}
\def\rmX{{\mathbf{X}}}
\def\rmY{{\mathbf{Y}}}
\def\rmZ{{\mathbf{Z}}}

\def\va{{\bm{a}}}

\def\vc{{\bm{c}}}

\def\vh{{\bm{h}}}

\def\vx{{\bm{x}}}
\def\vy{{\bm{y}}}
\def\vz{{\bm{z}}}

\def\mB{{\bm{B}}}

\def\mX{{\bm{X}}}
\def\mY{{\bm{Y}}}
\def\mZ{{\bm{Z}}}

\DeclareMathAlphabet{\mathsfit}{\encodingdefault}{\sfdefault}{m}{sl}
\SetMathAlphabet{\mathsfit}{bold}{\encodingdefault}{\sfdefault}{bx}{n}

\def\tZ{{\tens{Z}}}

\def\gX{{\mathcal{X}}}

\def\sR{{\mathbb{R}}}

\newcommand{\etens}[1]{\mathsfit{#1}}

\def\etC{{\etens{C}}}

\def\etH{{\etens{H}}}

\def\etW{{\etens{W}}}

\newcommand{\E}{\mathbb{E}}

\newcommand{\R}{\mathbb{R}}

\newcommand{\softmax}{\mathrm{softmax}}

\newcommand{\KL}{D_{\mathrm{KL}}}



%% file: encoder.bib
@article{Tishby2000,
  author       = {Naftali Tishby and
                  Fernando C. N. Pereira and
                  William Bialek},
  title        = {{The Information Bottleneck Method}},
  journal      = {CoRR},
  volume       = {physics/0004057},
  year         = {2000},
  url          = {http://arxiv.org/abs/physics/0004057},
  bibsource    = {dblp computer science bibliography, https://dblp.org}
}

@inproceedings{Kingma13,
  author       = {Diederik P. Kingma and
                  Max Welling},
  editor       = {Yoshua Bengio and
                  Yann LeCun},
  title        = {{Auto-Encoding Variational Bayes}},
  booktitle    = {2nd International Conference on Learning Representations, {ICLR} 2014,
                  Banff, AB, Canada, April 14-16, 2014, Conference Track Proceedings},
  year         = {2014},
  url          = {http://arxiv.org/abs/1312.6114},
  bibsource    = {dblp computer science bibliography, https://dblp.org}
}

@article{Kingma19,
  author       = {Diederik P. Kingma and
                  Max Welling},
  title        = {{An Introduction to Variational Autoencoders}},
  journal      = {CoRR},
  volume       = {abs/1906.02691},
  year         = {2019},
  url          = {http://arxiv.org/abs/1906.02691},
  eprinttype   = {arXiv},
  eprint       = {1906.02691},
  bibsource    = {dblp computer science bibliography, https://dblp.org}
}

@inproceedings{Durkan21,
  author       = {Yang Song and
                  Conor Durkan and
                  Iain Murray and
                  Stefano Ermon},
  editor       = {Marc'Aurelio Ranzato and
                  Alina Beygelzimer and
                  Yann N. Dauphin and
                  Percy Liang and
                  Jennifer Wortman Vaughan},
  title        = {{Maximum Likelihood Training of Score-Based Diffusion Models}},
  booktitle    = {Advances in Neural Information Processing Systems 34: Annual Conference
                  on Neural Information Processing Systems 2021, NeurIPS 2021, December
                  6-14, 2021, virtual},
  pages        = {1415--1428},
  year         = {2021},
  url          = {https://proceedings.neurips.cc/paper/2021/hash/0a9fdbb17feb6ccb7ec405cfb85222c4-Abstract.html},
  bibsource    = {dblp computer science bibliography, https://dblp.org}
}

@InProceedings{Celik25,
  title = 	 {{{DIME}: Diffusion-Based Maximum Entropy Reinforcement Learning}},
  author =   {Celik, Onur and Li, Zechu and Blessing, Denis and Li, Ge and Palenicek, Daniel and Peters, Jan and Chalvatzaki, Georgia and Neumann, Gerhard},
  booktitle = 	 {Proceedings of the 42nd International Conference on Machine Learning},
  pages = 	 {6958--6977},
  year = 	 {2025},
  editor = 	 {Singh, Aarti and Fazel, Maryam and Hsu, Daniel and Lacoste-Julien, Simon and Berkenkamp, Felix and Maharaj, Tegan and Wagstaff, Kiri and Zhu, Jerry},
  volume = 	 {267},
  series = 	 {Proceedings of Machine Learning Research},
  month = 	 {13--19 Jul},
  publisher =    {PMLR},
  url = 	 {https://proceedings.mlr.press/v267/celik25a.html},
}

@inproceedings{Chen18,
 author = {{Chen, Ricky T. Q. and Rubanova, Yulia and Bettencourt, Jesse and Duvenaud, David K}},
 booktitle = {Advances in Neural Information Processing Systems},
 editor = {S. Bengio and H. Wallach and H. Larochelle and K. Grauman and N. Cesa-Bianchi and R. Garnett},
 pages = {},
 publisher = {Curran Associates, Inc.},
 title = {Neural Ordinary Differential Equations},
 url = {https://proceedings.neurips.cc/paper_files/paper/2018/file/69386f6bb1dfed68692a24c8686939b9-Paper.pdf},
 volume = {31},
 year = {2018}
}

@InProceedings{Li20,
  title = 	 {Scalable Gradients and Variational Inference for
    Stochastic Differential Equations },
  author =       {Li, Xuechen and Wong, Ting-Kam Leonard and Chen, Ricky T. Q. and Duvenaud, David K.},
  booktitle = 	 {Proceedings of The 2nd Symposium on
    Advances in Approximate Bayesian Inference},
  pages = 	 {1--28},
  year = 	 {2020},
  editor = 	 {Zhang, Cheng and Ruiz, Francisco and Bui, Thang and Dieng, Adji Bousso and Liang, Dawen},
  volume = 	 {118},
  series = 	 {Proceedings of Machine Learning Research},
  month = 	 {08 Dec},
  publisher =    {PMLR},
  url = 	 {https://proceedings.mlr.press/v118/li20a.html}
}

@inproceedings{Premkumar25a,
    title={{Neural Entropy}},
    author={Akhil Premkumar},
    booktitle={The Thirty-ninth Annual Conference on Neural Information Processing Systems},
    year={2025},
    url={https://openreview.net/forum?id=f6AYwCvynr}
}

@inproceedings{Karras22,
  author       = {Tero Karras and
                  Miika Aittala and
                  Timo Aila and
                  Samuli Laine},
  editor       = {Sanmi Koyejo and
                  S. Mohamed and
                  A. Agarwal and
                  Danielle Belgrave and
                  K. Cho and
                  A. Oh},
  title        = {{Elucidating the Design Space of Diffusion-Based Generative Models}},
  booktitle    = {Advances in Neural Information Processing Systems 35: Annual Conference
                  on Neural Information Processing Systems 2022, NeurIPS 2022, New Orleans,
                  LA, USA, November 28 - December 9, 2022},
  year         = {2022},
  url          = {http://papers.nips.cc/paper\_files/paper/2022/hash/a98846e9d9cc01cfb87eb694d946ce6b-Abstract-Conference.html},
  bibsource    = {dblp computer science bibliography, https://dblp.org}
}

@article{Vincent11,
  title={A Connection between Score Matching and Denoising Autoencoders},
  author={Vincent, Pascal},
  journal={Neural computation},
  volume={23},
  number={7},
  pages={1661--1674},
  year={2011},
  publisher={MIT Press},
  doi={10.1162/NECO_a_00142}
}

@inproceedings{Franzese24,
  author    = {Giulio Franzese and Mustapha Bounoua and Pietro Michiardi},
  title     = {{MINDE}: Mutual Information Neural Diffusion Estimation},
  booktitle = {Proceedings of the International Conference on Learning Representations (ICLR)},
  year      = {2024},
  pages     = {16685--16716},
  url       = {https://proceedings.iclr.cc/paper_files/paper/2024/file/47f75e809409709c6d226ab5ca0c9703-Paper-Conference.pdf}
}

@inproceedings{Domingo25,
  author       = {Carles Domingo{-}Enrich and
                  Michal Drozdzal and
                  Brian Karrer and
                  Ricky T. Q. Chen},
  title        = {{Adjoint Matching: Fine-tuning Flow and Diffusion Generative Models
                  with Memoryless Stochastic Optimal Control}},
  booktitle    = {The Thirteenth International Conference on Learning Representations,
                  {ICLR} 2025, Singapore, April 24-28, 2025},
  publisher    = {OpenReview.net},
  year         = {2025},
  url          = {https://openreview.net/forum?id=xQBRrtQM8u},
  bibsource    = {dblp computer science bibliography, https://dblp.org}
}

@inproceedings{Domingo24,
  author       = {Carles Domingo{-}Enrich and
                  Jiequn Han and
                  Brandon Amos and
                  Joan Bruna and
                  Ricky T. Q. Chen},
  editor       = {Amir Globersons and
                  Lester Mackey and
                  Danielle Belgrave and
                  Angela Fan and
                  Ulrich Paquet and
                  Jakub M. Tomczak and
                  Cheng Zhang},
  title        = {{Stochastic Optimal Control Matching}},
  booktitle    = {Advances in Neural Information Processing Systems 38: Annual Conference
                  on Neural Information Processing Systems 2024, NeurIPS 2024, Vancouver,
                  BC, Canada, December 10 - 15, 2024},
  year         = {2024},
  url          = {http://papers.nips.cc/paper\_files/paper/2024/hash/cc32ec39a5073f61d38c338d963df30d-Abstract-Conference.html},
  bibsource    = {dblp computer science bibliography, https://dblp.org}
}

@inproceedings{Li26,
    title={{Q-Learning with Adjoint Matching}},
    author={Qiyang Li and Sergey Levine},
    booktitle={The Fourteenth International Conference on Learning Representations},
    year={2026},
    url={https://openreview.net/forum?id=vd4eNAdtO6}
}

@inproceedings{Premkumar26a,
    title={{On the Separability of Information in Diffusion Models}},
    author={Akhil Premkumar},
    booktitle={Forty-third International Conference on Machine Learning},
    year={2026},
    url={https://openreview.net/forum?id=Qc6OqkFAmO}
}

@inproceedings{Kong23,
  author       = {Xianghao Kong and
                  Rob Brekelmans and
                  Greg Ver Steeg},
  title        = {{Information-Theoretic Diffusion}},
  booktitle    = {The Eleventh International Conference on Learning Representations,
                  {ICLR} 2023, Kigali, Rwanda, May 1-5, 2023},
  publisher    = {OpenReview.net},
  year         = {2023},
  url          = {https://openreview.net/forum?id=UvmDCdSPDOW},
  bibsource    = {dblp computer science bibliography, https://dblp.org}
}

@inproceedings{Kong24,
  title     = {{Interpretable Diffusion via Information Decomposition}},
  author    = {Xianghao Kong and Ollie Liu and Han Li and Dani Yogatama and Greg Ver Steeg},
  booktitle = {Proceedings of the Twelfth International Conference on Learning Representations (ICLR)},
  year      = {2024},
  url       = {https://openreview.net/forum?id=X6tNkN6ate}
}

@inproceedings{Haarnoja18,
  author       = {Tuomas Haarnoja and
                  Aurick Zhou and
                  Pieter Abbeel and
                  Sergey Levine},
  editor       = {Jennifer G. Dy and
                  Andreas Krause},
  title        = {{Soft Actor-Critic: Off-Policy Maximum Entropy Deep Reinforcement Learning
                  with a Stochastic Actor}},
  booktitle    = {Proceedings of the 35th International Conference on Machine Learning,
                  {ICML} 2018, Stockholmsm{\"{a}}ssan, Stockholm, Sweden, July
                  10-15, 2018},
  series       = {Proceedings of Machine Learning Research},
  pages        = {1856--1865},
  publisher    = {{PMLR}},
  year         = {2018},
  url          = {http://proceedings.mlr.press/v80/haarnoja18b.html},
  bibsource    = {dblp computer science bibliography, https://dblp.org}
}

@InProceedings{Cremer18,
  title = 	 {{Inference Suboptimality in Variational Autoencoders}},
  author =   {Cremer, Chris and Li, Xuechen and Duvenaud, David},
  booktitle ={Proceedings of the 35th International Conference on Machine Learning},
  pages = 	 {1078--1086},
  year = 	 {2018},
  editor = 	 {Dy, Jennifer and Krause, Andreas},
  volume = 	 {80},
  series = 	 {Proceedings of Machine Learning Research},
  month = 	 {10--15 Jul},
  publisher ={PMLR},
  url = 	 {https://proceedings.mlr.press/v80/cremer18a.html}
}

@inproceedings{Kim18,
  title     = {{Semi-Amortized Variational Autoencoders}},
  author    = {Kim, Yoon and Wiseman, Sam and Miller, Andrew and Sontag, David and Rush, Alexander},
  booktitle = {Proceedings of the 35th International Conference on Machine Learning},
  pages     = {2678--2687},
  year      = {2018},
  volume    = {80},
  series    = {Proceedings of Machine Learning Research},
  publisher = {PMLR},
}

@inproceedings{Higgins17,
    title     = {{beta-{VAE}: Learning Basic Visual Concepts
                 with a Constrained Variational Framework}},
    author    = {Higgins, Irina and
                 Matthey, Loic and
                 Pal, Arka and
                 Burgess, Christopher and
                 Glorot, Xavier and
                 Botvinick, Matthew and
                 Mohamed, Shakir and
                 Lerchner, Alexander},
    booktitle = {International Conference on Learning Representations},
    year      = {2017},
    url       = {https://openreview.net/forum?id=Sy2fzU9gl}
}

@article{Dempster77,
    title     = {{Maximum Likelihood from Incomplete Data via the {EM} Algorithm}},
    author    = {Dempster, Arthur P. and
                 Laird, Nan M. and
                 Rubin, Donald B.},
    journal   = {Journal of the Royal Statistical Society, Series B},
    volume    = {39},
    number    = {1},
    pages     = {1--38},
    year      = {1977}
}

@InProceedings{Rezende15,
  title = 	 {{Variational Inference with Normalizing Flows}},
  author = 	 {Rezende, Danilo and Mohamed, Shakir},
  booktitle = 	 {Proceedings of the 32nd International Conference on Machine Learning},
  pages = 	 {1530--1538},
  year = 	 {2015},
  editor = 	 {Bach, Francis and Blei, David},
  volume = 	 {37},
  series = 	 {Proceedings of Machine Learning Research},
  address = 	 {Lille, France},
  month = 	 {07--09 Jul},
  publisher =    {PMLR},
  url = 	 {https://proceedings.mlr.press/v37/rezende15.html}
}

@inproceedings{Lipman23,
  author       = {Yaron Lipman and
                  Ricky T. Q. Chen and
                  Heli Ben{-}Hamu and
                  Maximilian Nickel and
                  Matthew Le},
  title        = {{Flow Matching for Generative Modeling}},
  booktitle    = {The Eleventh International Conference on Learning Representations,
                  {ICLR} 2023, Kigali, Rwanda, May 1-5, 2023},
  publisher    = {OpenReview.net},
  year         = {2023},
  url          = {https://openreview.net/forum?id=PqvMRDCJT9t},
  bibsource    = {dblp computer science bibliography, https://dblp.org}
}

@inproceedings{Lucas19,
  title     = {{Don't Blame the ELBO! A Linear VAE Perspective on Posterior Collapse}},
  author    = {Lucas, James and Tucker, George and Grosse, Roger B. and Norouzi, Mohammad},
  booktitle = {Advances in Neural Information Processing Systems (NeurIPS)},
  volume    = {32},
  year      = {2019},
  url       = {https://neurips.cc}
}

@inproceedings{He19,
  title     = {{Lagging Inference Networks and Posterior Collapse in Variational Autoencoders}},
  author    = {He, Junxian and Spokoyny, Daniel and Neubig, Graham and Berg-Kirkpatrick, Taylor},
  booktitle = {International Conference on Learning Representations (ICLR)},
  year      = {2019},
  url       = {https://openreview.net/forum?id=rylDfnCqF7}
}

@InProceedings{Alemi18,
  title = 	 {{Fixing a Broken ELBO}},
  author =   {Alemi, Alexander and Poole, Ben and Fischer, Ian and Dillon, Joshua and Saurous, Rif A. and Murphy, Kevin},
  booktitle = 	 {Proceedings of the 35th International Conference on Machine Learning},
  pages = 	 {159--168},
  year = 	 {2018},
  editor = 	 {Dy, Jennifer and Krause, Andreas},
  volume = 	 {80},
  series = 	 {Proceedings of Machine Learning Research},
  month = 	 {10--15 Jul},
  publisher =    {PMLR},
  url = 	 {https://proceedings.mlr.press/v80/alemi18a.html}
}

@inproceedings{Song21,
  author       = {Yang Song and
                  Jascha Sohl{-}Dickstein and
                  Diederik P. Kingma and
                  Abhishek Kumar and
                  Stefano Ermon and
                  Ben Poole},
  title        = {{Score-Based Generative Modeling through Stochastic Differential Equations}},
  booktitle    = {9th International Conference on Learning Representations, {ICLR} 2021,
                  Virtual Event, Austria, May 3-7, 2021},
  publisher    = {OpenReview.net},
  year         = {2021},
  url          = {https://openreview.net/forum?id=PxTIG12RRHS},
  bibsource    = {dblp computer science bibliography, https://dblp.org}
}

@inproceedings{Ho20,
  author       = {Jonathan Ho and
                  Ajay Jain and
                  Pieter Abbeel},
  editor       = {Hugo Larochelle and
                  Marc'Aurelio Ranzato and
                  Raia Hadsell and
                  Maria{-}Florina Balcan and
                  Hsuan{-}Tien Lin},
  title        = {Denoising Diffusion Probabilistic Models},
  booktitle    = {Advances in Neural Information Processing Systems 33: Annual Conference
                  on Neural Information Processing Systems 2020, NeurIPS 2020, December
                  6-12, 2020, virtual},
  year         = {2020},
  url          = {https://proceedings.neurips.cc/paper/2020/hash/4c5bcfec8584af0d967f1ab10179ca4b-Abstract.html},
  bibsource    = {dblp computer science bibliography, https://dblp.org}
}

@inproceedings{Dhariwal2021,
 author = {Dhariwal, Prafulla and Nichol, Alexander},
 booktitle = {Advances in Neural Information Processing Systems},
 editor = {M. Ranzato and A. Beygelzimer and Y. Dauphin and P.S. Liang and J. Wortman Vaughan},
 pages = {8780--8794},
 publisher = {Curran Associates, Inc.},
 title = {Diffusion Models Beat GANs on Image Synthesis},
 url = {https://proceedings.neurips.cc/paper_files/paper/2021/file/49ad23d1ec9fa4bd8d77d02681df5cfa-Paper.pdf},
 volume = {34},
 year = {2021}
}

@inproceedings{Vaswani17,
    title     = {Attention is All you Need},
    author    = {Vaswani, Ashish and
                 Shazeer, Noam and
                 Parmar, Niki and
                 Uszkoreit, Jakob and
                 Jones, Llion and
                 Gomez, Aidan N. and
                 Kaiser, Lukasz and
                 Polosukhin, Illia},
    booktitle = {Advances in Neural Information Processing Systems},
    volume    = {30},
    year      = {2017}
}

@inproceedings{Tancik20,
    title     = {Fourier Features Let Networks Learn High Frequency Functions
                 in Low Dimensional Domains},
    author    = {Tancik, Matthew and
                 Srinivasan, Pratul and
                 Mildenhall, Ben and
                 Fridovich-Keil, Sara and
                 Raghavan, Nithin and
                 Singhal, Utkarsh and
                 Ramamoorthi, Ravi and
                 Barron, Jonathan and
                 Ng, Ren},
    booktitle = {Advances in Neural Information Processing Systems},
    volume    = {33},
    pages     = {7537--7547},
    year      = {2020}
}

@article{Peebles22,
  author       = {William Peebles and
                  Saining Xie},
  title        = {Scalable Diffusion Models with Transformers},
  journal      = {CoRR},
  volume       = {abs/2212.09748},
  year         = {2022},
  url          = {https://doi.org/10.48550/arXiv.2212.09748},
  doi          = {10.48550/arxiv.2212.09748},
  eprinttype   = {arXiv},
  eprint       = {2212.09748},
  bibsource    = {dblp computer science bibliography, https://dblp.org}
}

@article{Maoutsa20,
    title   = {Interacting Particle Solutions of {Fokker--Planck} Equations
               Through Gradient--Log--Density Estimation},
    author  = {Maoutsa, Dimitra and
               Reich, Sebastian and
               Opper, Manfred},
    journal = {Entropy},
    volume  = {22},
    number  = {8},
    pages   = {802},
    year    = {2020},
    url     = {https://www.mdpi.com/1099-4300/22/8/802}
}

@misc{JAX18,
  author = {James Bradbury and Roy Frostig and Peter Hawkins and Matthew James Johnson and Chris Leary and Dougal Maclaurin and George Necula and Adam Paszke and Jake Vander{P}las and Skye Wanderman-{M}ilne and Qiao Zhang},
  title = {{JAX}: composable transformations of {P}ython+{N}um{P}y programs},
  url = {http://github.com/jax-ml/jax},
  version = {0.3.13},
  year = {2018},
}

@article{Shamir2010,
  title   = {{Learning and Generalization with the Information Bottleneck}},
  author  = {Shamir, Ohad and Sabato, Sivan and Tishby, Naftali},
  journal = {Theoretical Computer Science},
  volume  = {411},
  number  = {29},
  pages   = {2696--2711},
  year    = {2010},
  doi     = {10.1016/j.tcs.2010.04.006},
  note    = {Algorithmic Learning Theory (ALT 2008)}
}

@article{LeCun98,
  author    = {LeCun, Yann and Bottou, Léon and Bengio, Yoshua and Haffner, Patrick},
  title     = {Gradient-Based Learning Applied to Document Recognition},
  journal   = {Proceedings of the IEEE},
  year      = {1998},
  volume    = {86},
  number    = {11},
  pages     = {2278--2324},
  doi       = {10.1109/5.726791}
}

@techreport{Krizhevsky09,
  title={{Learning multiple layers of features from tiny images}},
  author={Krizhevsky, Alex},
  year={2009},
  institution={University of Toronto},
  note={\url{https://www.cs.toronto.edu/~kriz/learning-features-2009-TR.pdf}}
}

@inproceedings{Deng09,
  author    = {Deng, Jia and Dong, Wei and Socher, Richard and Li, Li-Jia and Li, Kai and Fei-Fei, Li},
  title     = {{ImageNet}: A Large-Scale Hierarchical Image Database},
  booktitle = {2009 IEEE Conference on Computer Vision and Pattern Recognition (CVPR)},
  year      = {2009},
  pages     = {248--255},
  doi       = {10.1109/CVPR.2009.5206848}
}

@inproceedings{Liu2015,
  title     = {Deep Learning Face Attributes in the Wild},
  author    = {Liu, Ziwei and Luo, Ping and Wang, Xiaogang and Tang, Xiaoou},
  booktitle = {Proceedings of International Conference on Computer Vision (ICCV)},
  month     = {December},
  year  = {2015} 
}

@article{Karras17,
  author       = {Tero Karras and
                  Timo Aila and
                  Samuli Laine and
                  Jaakko Lehtinen},
  title        = {Progressive Growing of GANs for Improved Quality, Stability, and Variation},
  journal      = {CoRR},
  volume       = {abs/1710.10196},
  year         = {2017},
  url          = {http://arxiv.org/abs/1710.10196},
  eprinttype   = {arXiv},
  eprint       = {1710.10196},
  bibsource    = {dblp computer science bibliography, https://dblp.org}
}
